\documentclass[10pt,twocolumn,letterpaper]{article}
\date{June 2026}
\usepackage[margin=0.75in,columnsep=0.25in]{geometry}

\usepackage{times}
\usepackage[utf8]{inputenc}
\usepackage[T1]{fontenc}

\usepackage{hyperref}
\usepackage{url}
\usepackage{booktabs}
\usepackage{multirow}
\usepackage{amsfonts}
\usepackage{amsmath}
\usepackage{amssymb}
\usepackage{nicefrac}
\usepackage{microtype}
\usepackage{xcolor}
\usepackage{graphicx}
\usepackage{algorithm}
\usepackage{algorithmic}
\usepackage{subcaption}
\usepackage{titlesec}
\usepackage[numbers]{natbib}
\usepackage{caption}

\titleformat{\section}{\normalfont\large\bfseries}{\thesection}{1em}{}
\titleformat{\subsection}{\normalfont\normalsize\bfseries}{\thesubsection}{1em}{}
\titleformat{\subsubsection}{\normalfont\normalsize\itshape}{\thesubsubsection}{1em}{}
\titlespacing*{\section}{0pt}{2ex plus 1ex minus .2ex}{1ex plus .2ex}
\titlespacing*{\subsection}{0pt}{1.5ex plus 1ex minus .2ex}{0.5ex plus .2ex}
\titlespacing*{\subsubsection}{0pt}{1ex plus 1ex minus .2ex}{0.5ex plus .2ex}

\usepackage{xurl}

\graphicspath{{media/}{figures/out/}}

\title{Learning to Fold: prizewinning solution \\
at LeHome Challenge 2026 (1st place online, 2nd offline)}

\author{%
  \begin{tabular}[t]{@{}c@{}}
    Ilia Larchenko \\
    \texttt{\small ilya.larchenko@gmail.com}
  \end{tabular}
  \vspace{5.0mm}
  \\[1.0ex]
  Independent Researcher
}

\begin{document}

\twocolumn[
  \begin{@twocolumnfalse}
    \maketitle
    \begin{abstract}
    I describe my solution to the \textbf{LeHome Challenge 2026}, an ICRA 2026 competition on bimanual garment folding. The system placed \textbf{1st of 62 teams in the online (simulation) round} and \textbf{2nd in the real-world final}. It improves a vision-language-action (VLA) policy with a reinforcement-learning loop. The policy is its own value function: the same network that predicts actions also predicts success, progress, and a few task-relevant future quantities, and those predictions drive advantage estimation, live failure detection, and candidate selection.

    The work mostly recombines existing RL ideas with engineering and optimization contributions that can be used together as one recipe or individually:

    \begin{itemize}
        \item AWR + RECAP combined for flow-matching VLA;
        \item an asynchronous distributed training / rollout pipeline through HuggingFace Hub;
        \item inference-time hyperparameters optimization via Thompson sampling;
        \item a sim-to-real recipe with camera-alignment tooling, heavy augmentation and DAgger-like HIL data collection.
    \end{itemize}

    This report is an engineering case study, not a controlled experiment: the system was built iteratively under competition pressure with little formal ablations, so I describe what I did and what shipped, not which pieces were necessary.
    \end{abstract}
    \vspace{1em}
  \end{@twocolumnfalse}
]

\section{Introduction}

\subsection{The LeHome Challenge 2026}

\begin{figure*}[!htb]
  \centering
  \includegraphics[width=0.9\textwidth]{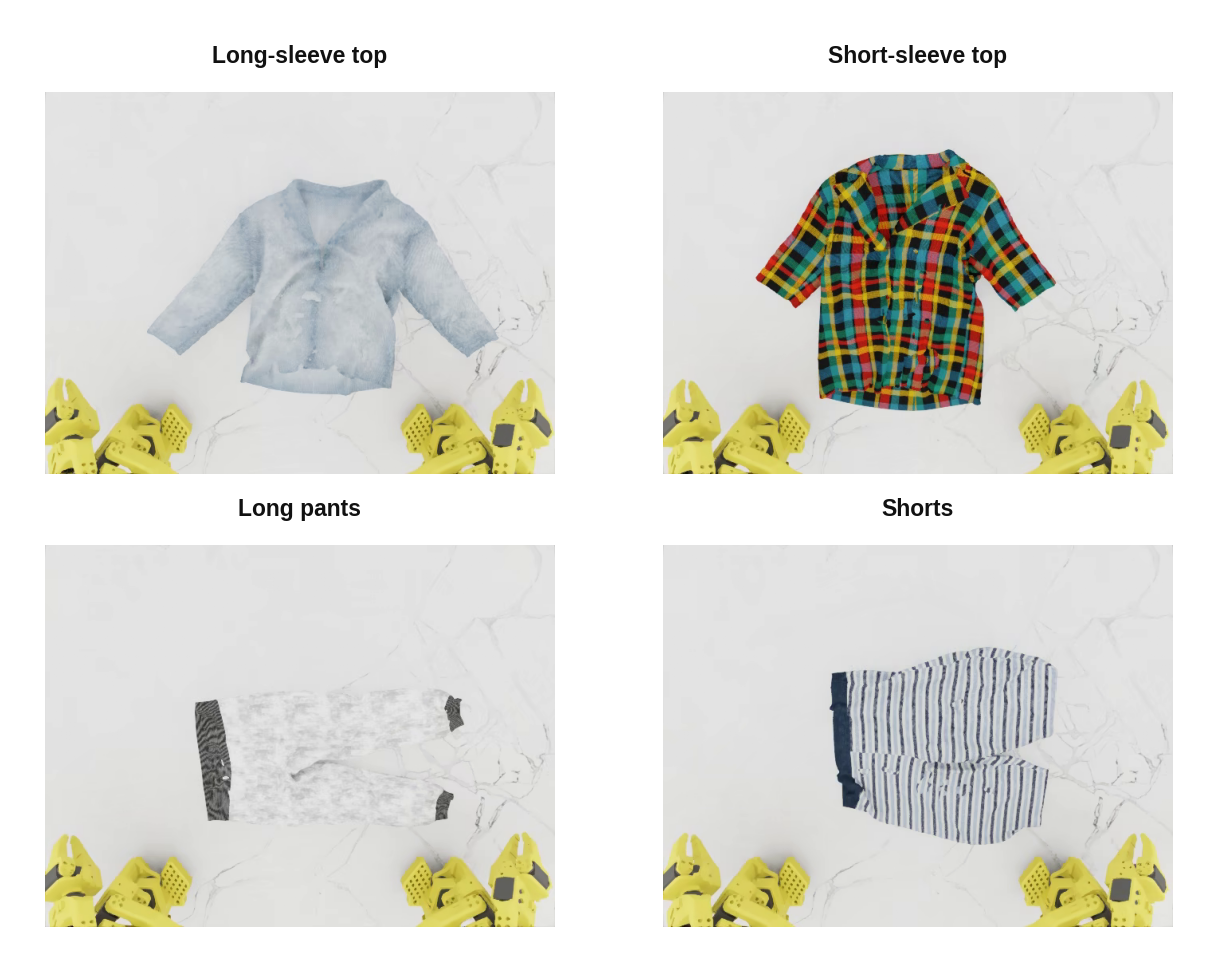}
  \caption{The four garment types, seen from the overhead camera in the original simulation behavior-cloning dataset released by the organizers.}
  \label{fig:sim_garment_types}
\end{figure*}

The LeHome Challenge 2026~\cite{lehome2026, li2026lehome} is the simulation-driven robotics competition centered on \textbf{deformable-object manipulation}. The task is garment folding: fold a single garment lying flat on a table, using a \textbf{bimanual SO-ARM101} setup~\cite{soarm100, lerobot2024}--- two 6-DOF arms, a 12-dimensional joint action space at 30 Hz in sim (20 Hz in the real round), and three RGB cameras (one overhead, one on each wrist). Depth is available from the overhead camera but I did not use it. Four garment types are evaluated: long-sleeved tops, short-sleeved tops, long pants, and shorts (Figure~\ref{fig:sim_garment_types}).

Success is binary and defined geometrically. Each garment carries a set of keypoints, and success is a combination of pairwise keypoint-distance conditions: pairs that should meet (e.g. two sleeves) must come closer than a threshold, pairs that should stay apart must remain farther away. There are 5 conditions for tops and 4 for pants. I reuse these same conditions to build dense intermediate rewards (Section~\ref{sec:reward}).

The competition ran in two phases:

\begin{itemize}
    \item \textbf{Online round (simulation), Feb--Apr 2026.} Open to all teams, with continuous submissions evaluated in Isaac Sim~\cite{isaacsim2026} and ranked on a public leaderboard by overall success rate. There is no partial credit --- a fold passes all conditions or it fails. Each garment type is scored over 20 instances: 10 \textbf{seen} garments, for which the organizers released a scripted-policy behavior-cloning dataset, and 10 \textbf{unseen}. Of the unseen, 2 per type were public (completely accessible during development, but with no organizer training data) and 8 per type were private, never exposed at all. Sections~\ref{sec:training}--\ref{sec:results} cover this round.
    \item \textbf{Real-world final, June 2026 at ICRA (Vienna).} The top 8 simulation teams were evaluated on-site on a physical robot, as part of the ICRA 2026 conference. The offline evaluation used slightly different success criteria, awarding points for partial success (e.g. folding a single sleeve), with an organizers' jury scoring the teams' results. Each type was evaluated on 5 garments --- 3 seen (provided in the organizer BC dataset) and 2 unseen. Section~\ref{sec:sim2real} covers this round.
\end{itemize}

An important protocol detail: \textbf{the garment category is not given at evaluation time} --- garments are loaded at random and the policy cannot read the label, so a policy that needs the garment type must infer it. I handle this with a learned garment-type input token and an inference-time classifier bootstrap (\S4.2, \S7.6).

The challenge ran in NVIDIA Isaac Lab (built on Isaac Sim) on the organizers' released environment, assets, and success checker~\cite{lehome2026, li2026lehome}.

\subsection{Key challenges}

The task combines several difficulties, and most of the system exists to address one of them:

\begin{enumerate}
    \item \textbf{Cloth is deformable and hard to manipulate.} Small differences in the trajectory can lead to different garment states. Plain behavior cloning on the provided scripted demonstrations is not very robust as the expert trajectories are clean and inflexible. \textit{(Motivates RL, Section~\ref{sec:training}, and recovery data, \S3.4 / \S9.10.)}
    \item \textbf{The reward is sparse and binary.} Nothing is observable until the episode ends, and an easy success looks identical to a hard one. All intermediate signal has to be engineered. \textit{(Section~\ref{sec:reward}.)}
    \item \textbf{Generalization to unseen garments.} Significant part of the leaderboard set is unseen with no training data at all. Heavy domain randomization in both the sim (\S\ref{sec:env_aug}) and real (\S\ref{sec:heavy_aug}) rounds is the main lever I used against this.
    \item \textbf{No access to the evaluation robot.} For the real round I never had the actual evaluation rig, so transfer was really sim $\to$ my robot $\to$ their robot, with an extra generalization step baked in. \textit{(\S9.1.)}
\end{enumerate}

\subsection{Approach summary}

This is a map of the rest of the report, in reading order.

\textbf{RL training, the flywheel (Section~\ref{sec:training}).} The system is an \textbf{asynchronous loop} of three independent components that talk only through HuggingFace Hub: a training worker, any number of rollout workers, and a manual DAgger station. Advantage drives the policy two ways at once --- AWR~\cite{peng2019awr} through the \textit{sampler} (high-advantage frames are loaded more often) and RECAP-style~\cite{recap2025} advantage \textit{conditioning} (advantage as an input, which unlocks classifier-free guidance at inference). I argue this conditioning/reweighting family suits flow-matching VLAs better than PPO-style methods. Training ran on a single H200; rollouts were collected mostly on RTX PRO 6000 GPU.

\textbf{Data collection (Section~\ref{sec:data}).} Everything built around the simulator: parallel sims behind one stateless policy server, multiple rollout-collection strategies, success/failure physics-state snapshots for replay and hard-mining, an environment-augmentation engine, and the asynchronous DAgger loop for hard cases.

\textbf{Policy architecture (Section~\ref{sec:architecture}).} I start from the policy my team built to win the BEHAVIOR-1K Challenge 2025~\cite{behavior1k2025}, that in turn is an extension of $\pi_{0.5}$~\cite{black2025pi05}: a frozen SigLIP encoder~\cite{zhai2023siglip}, a Gemma-2B prefix transformer~\cite{gemma2024}, and a Gemma-300M action expert that emits 30-step, 12-dimensional action chunks by flow matching. Carried over from that work (not contributions here): no language input, correlated flow-matching noise, soft inpainting between chunks, and cross-layer KV-cache mixing. My additions for LeHome are auxiliary heads as part of the main model (\S5), a garment-type input token (\S4.2), advantage conditioning (\S4.3), multi-signal AdaRMS conditioning (\S4.4), exclusive self-attention (\S4.5), and smooth per-timestep action normalization (\S4.6).

\textbf{The policy as its own value function (Section~\ref{sec:auxheads}).} A single learned query token feeds a set of cheap linear heads that all read from the image tokens only: success probability, task completion, garment type, keypoints distances and --- 30 frames ahead --- future keypoints distances and an action-conditional success residual that acts as a Q-function. Keeping value, Q, and a cheap world-model substitute inside the policy means one model to train and serve, and lets these signals share representation with the action head.

\textbf{Reward and advantage (Section~\ref{sec:reward}).} The binary success is densified into per-garment checkpoints built from the challenge's own keypoint conditions, with all reward withdrawn on failure so the episode return stays binary. The success head supplies a dampened (CUPED-style) value baseline, a completion head supplies a progress signal, and the two are combined with GAE into per-frame advantages that degrade gracefully toward outcome-only baselines as rollouts go stale.

\textbf{Inference-time optimization (Section~\ref{sec:inference}).} The same checkpoint can behave differently at inference time depending on how it is run. I tune execution length, playback speed, inpainting onset, guidance scale, noise temperature, and best-of-N candidate count \textbf{per garment type}, found cheaply online with a Thompson-sampling bandit during rollout collection.

\textbf{Online-round results (Section~\ref{sec:results}).} 1st of 62 teams at 79.63\% overall success, ahead of second place by 6.1 points.

\textbf{Sim-to-real (Section~\ref{sec:sim2real}).} A one-week sprint: start from a late-but-not-latest sim checkpoint, strip out the sim-only machinery, and fine-tune on a mix of organizer data, my own teleop/DAgger, and augmented sim replays --- with heavy augmentation, motion-velocity alignment, and a camera-overlay calibration tool. Result: 2nd in the real-world final.

\textbf{Discussion (Section~\ref{sec:discussion}).} What I would keep, what was hard, the unexpected robustness to setup changes, the open problem of autonomous exploration/recovery, and why fusing the two rounds approaches into a single pipeline should do much better.

\subsection{Related work}

The benchmark and simulator are the organizers' own~\cite{lehome2026, li2026lehome}. On the policy side I build directly on \textbf{$\pi_{0.5}$}~\cite{black2025pi05} --- a VLA with a flow-matching action expert --- through my team's \textbf{BEHAVIOR-1K} solution~\cite{behavior1k2025}, which extended it with the carried-over components above and which I reuse without changes. For improving a VLA with RL I rely on \textbf{AWR}~\cite{peng2019awr} (and similar advantage-weighted regression methods such as AWAC~\cite{nair2020awac}) and \textbf{RECAP}-style advantage conditioning~\cite{recap2025}, the latter from Physical Intelligence's $\pi^{*}_{0.6}$ work. The human-in-the-loop component follows \textbf{DAgger}~\cite{ross2011dagger} and related interactive approaches such as HIL-SERL~\cite{hilserl2024} with the goal of learning from online human corrections.

\section{RL Training}
\label{sec:training}

\subsection{The flywheel}

The whole system is an asynchronous loop of three independent components that communicate only through HuggingFace Hub (Figure~\ref{fig:system_overview}):

\begin{figure*}[!htb]
  \centering
  \includegraphics[width=0.92\textwidth]{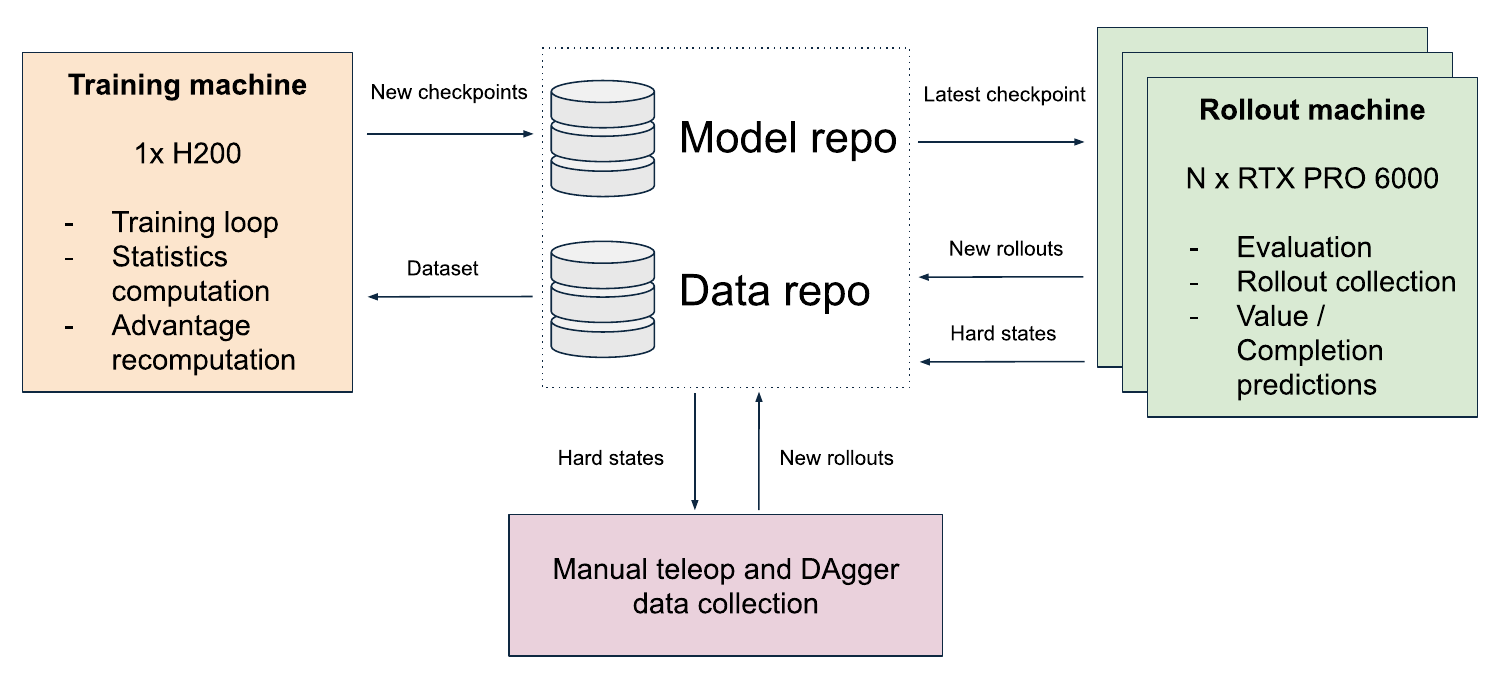}
  \caption{System overview: the asynchronous training / rollout / DAgger flywheel. The three components share state only through the HuggingFace Hub.}
  \label{fig:system_overview}
\end{figure*}

\begin{itemize}
    \item \textbf{Training worker} (one GPU machine): trains continuously, recomputes advantages across all rollout datasets before each iteration, uploads checkpoints.
    \item \textbf{Rollout workers} (any number of machines): pulls the latest checkpoint from the Hub, runs 3--5 parallel Isaac Sim instances each, uploads collected episodes with the values predicted during collection. Because rollout machines only need the Hub, scaling data collection is just starting another machine.
    \item \textbf{Manual DAgger station}: a human fixes saved failure states via teleop; the resulting episodes ship through the same Hub channel (\S3.4).
\end{itemize}

There are no synchronization barriers: the trainer trains on whatever data has arrived, the workers collect with whatever checkpoint is newest. A background HF-sync daemon on each machine handles uploads/downloads so neither training nor collection ever blocks on the network. One iteration of the trainer loop is: download new rollouts $\to$ recompute advantages over \textit{all} rollout datasets $\to$ train $\sim$1000 steps $\to$ upload checkpoint every $\sim$500 steps.

\subsection{Why AWR + RECAP}

If you have a BC-pretrained policy with a non-zero success rate, there are two ways to make it better: teach it to complete the task more cleanly on the first try, or teach it to recover from failures. Ideally you do both; in practice I mostly did the first in the first round of the competition and the second in the second round --- not entirely intentionally. AWR + RECAP are very good for the first goal. The initial BC data was very clean --- no failures and no recoveries --- which makes it a great base to refine but a poor source of recovery behavior. My bet for recovery was DAgger, but it turned out to be hard in simulation and didn't help much (\S3.4); success replays and augmentations made the policy more robust and reliable, but robustness is not the same as \textit{recovering} a ruined state.

Why this family of algorithms at all? Mainstream online RL --- PPO~\cite{schulman2017ppo} and its group-relative variant GRPO~\cite{shao2024grpo} --- is built around log-probability policy-gradient updates and doesn't transfer cleanly to flow-matching VLAs. Though there are attempts to adapt PPO-like logic to flow-matching VLAs, some with promising results.

Another problem is that valid actions occupy a tiny manifold inside the prediction space, and any algorithm that ``discourages bad actions'' --- pushing probability \textit{away} from something --- mostly pushes predictions off that manifold.

Conditioning and reweighting methods, in contrast, never leave the manifold: they only redistribute probability mass \textit{toward} good actions that the policy already produces. Their weakness is the flip side --- very limited exploration and discovery. For this competition that trade-off is favorable: I need to reliably fold the garment in one try, not to discover qualitatively new behaviors.

My preference for the conditioning/reweighting family is a subjective bet, that intuitively should work better for this problem.

So the training signal is consumed in two complementary ways:

\begin{itemize}
    \item \textbf{AWR}~\cite{peng2019awr}: high-advantage frames are trained on more often (\S2.3) --- the model eventually behaves better than its average rollout.
    \item \textbf{RECAP-style conditioning}~\cite{recap2025}: it feeds the advantage in as a conditioning \textit{input} (\S4.3), telling the model to ``predict good actions only'' --- which also unlocks classifier-free guidance (CFG) at inference.
\end{itemize}

Each of the two is a proven standalone RL approach; they rely on the same primitives, complement each other's strengths, and combine well.

Figure~\ref{fig:awr_recap} is a toy picture of what the two mechanisms do to the action distribution --- behavior data as a mixture of a large ``bad actions'' mode and a smaller ``good actions'' mode:

\begin{figure*}[!htb]
  \centering
  \includegraphics[width=0.9\textwidth]{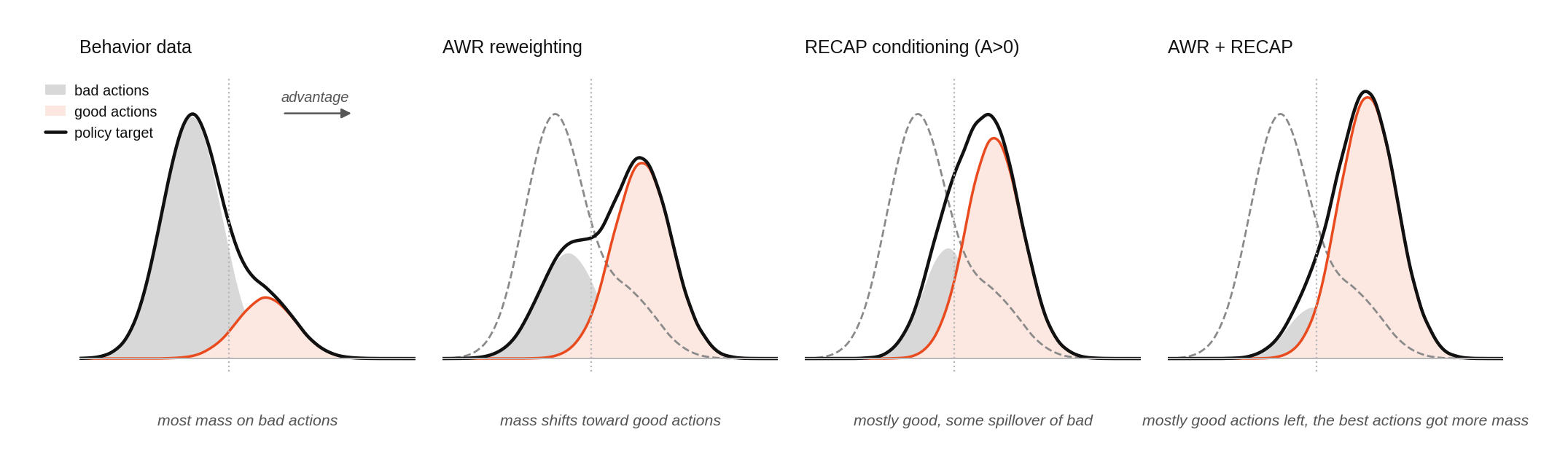}
  \caption{AWR + RECAP toy illustration.}
  \label{fig:awr_recap}
\end{figure*}

AWR reweights the mixture toward the good mode; RECAP conditioning selects the positive-advantage slice (mostly good, with some spillover of bad); doing both simultaneously leaves the policy target almost entirely on the good mode.

\subsection{AWR through the sampler, not the loss}

Advantage Weighted Regression (AWR)~\cite{peng2019awr} belongs to the regression family of RL algorithms, alongside similar advantage-weighted methods such as AWAC~\cite{nair2020awac}: instead of a policy-gradient update, the actor is trained by supervised regression onto the actions in the data, with each sample weighted by $e^{A/\beta}$. The original paper derives this as a solution to constrained policy improvement --- maximize expected improvement while staying close (in KL) to the sampling policy.

Contrary to the original formulation, I apply the weights through \textit{sampling} instead of loss weighting --- equivalent in expectation, but more data-efficient:

\[ P(\text{sample frame } i) \;\propto\; e^{\mathrm{clip}(A_i,\,-2,\,2)}. \]

The flow-matching loss over the batch is plain unweighted MSE. A frame with weight $e^{-2}$ is not down-weighted after being loaded --- it is simply almost never loaded, so its images are never decoded and never occupy batch slots. Effective batch utilization stays at 100\% of the weight mass, and most of the compute is spent on the good actions we actually want to learn.

One correction is needed: the auxiliary heads (\S\ref{sec:auxheads}) (success, completion, garment type, \ldots) must stay \textit{unbiased} --- their targets are statistics of the data distribution, not of the advantage-tilted distribution. Every sampled frame therefore carries an importance weight

\[ w_i = \frac{1}{N \, p_i \, T_{\mathrm{ep}(i)}}, \]

(inverse sampling (IS) probability, normalized per episode length), and all auxiliary losses are weighted by $w_i$ while the action loss ignores it. The similar machinery is applied to BC and DAgger sources: there the per-frame priority is failure-rate-proportional, $P \propto e^{3(1-\mathrm{SR}_{\mathrm{garment}})}$, so garments the policy still struggles with are over-sampled, and the IS weights again debias the aux heads.

\subsection{Runtime multi-dataset sampling}

Datasets are never merged. The loader holds every source --- the BC dataset, all DAgger sessions, and every RL rollout batch collected so far --- and samples among them at run time according to per-source shares. This makes the data mix a \textit{config parameter} rather than a preprocessing step: shares change every iteration without any extra processing.

\begin{itemize}
    \item \textbf{RL rollout datasets} decay by $0.98$ per training iteration (floored at $0.1$, to drop overly stale data and save disk space): fresh on-policy data dominates, old data fades. Over time a dataset's advantages also partially switch to the outcome-only segment form (\S6.5).
    \item \textbf{BC dataset} keeps a fixed sampling rate, so its effective weight decreases toward an asymptote as the total dataset grows. The sample rate within the BC dataset depends on the overall garment success rate --- harder garments are sampled more.
    \item Some \textbf{old successful rollouts} whose datasets had already decayed out of the schedule were kept in the mix at a higher weight --- mostly to retain successful examples for the garments with the lowest success rates.
\end{itemize}

\subsection{Checkpoint rollbacks}

A practical trick I found very useful: train and collect data for a while, then \textbf{roll back to a checkpoint from a few days ago and continue training it on all the data collected since} --- including the data collected by the newer checkpoints. The rolled-back model sees a large batch of fresh, diverse, partially off-policy experience at once instead of having co-evolved with it, which reliably kicked the policy out of local optima that the continuous training had settled into. $\pi^{*}_{0.6}$~\cite{recap2025} actually does the same systematically --- every RL training iteration restarts from the same baseline checkpoint. I did it less methodically: 3 rollbacks during the first round and 1 during the second.

\subsection{Training setup}

I did most of the training on a single H200 machine with batch size 192 --- around 300k steps in total plus some ad-hoc experiments. Rollout collection ran mostly on separate RTX PRO 6000 GPU (the same as the DAgger workstation, \S9.10). The recipe itself is simple: $\pi_{0.5}$ base weights, a 20k-step BC warm-up before the first rollouts, cosine LR ($10^{-4} \to 10^{-5}$ over 100k steps), AdamW with the extra aux-head weight decay (\S5.3), bfloat16, frozen SigLIP, 5 flow samples per batch item. Light training-time image augmentations (color jitter, blur, top-camera crop/rotate) were on at all times; the sim-to-real phase later made them much more aggressive (see \S\ref{sec:sim2real}).

\section{Data Collection}
\label{sec:data}

Simulation speed is the bottleneck of the whole system --- an episode takes about 30 s in Isaac Sim --- so the collection side is built around squeezing useful episodes out of every sim-hour.

\subsection{Plumbing}

\begin{figure}[!htb]
  \centering
  \includegraphics[width=\linewidth]{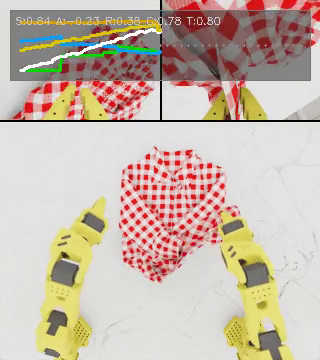}
  \caption{A mid-episode frame from a rollout debug video. The three camera views combined. The overlay shows the per-frame success probability ($S$), advantage ($A$), reward ($R$), completion ($C$), and time-to-completion ($T$), with their traces over the episode.}
  \label{fig:debug_overlay}
\end{figure}

The collection stack went through a lot of engineering optimization; I won't cover it in detail here, but the full implementation is released with the code. The points that mattered most:

\begin{itemize}
    \item \textbf{3--5 sim processes per machine} (the competition environment doesn't support multiple scenes in one process), each a thin client to one shared, stateless policy server; action chunks are cached client-side.
    \item \textbf{Early termination and recovery}: stop immediately on success; a stuck detector (state, action, and predicted-value variance) trims hopeless episodes; a watchdog restarts hung sims.
    \item \textbf{Everything recorded at collection time}: episodes are saved with the values predicted during the rollout (consumed by advantage computation, \S6.5), plus a debug video with the reward/value/advantage overlay (Figure~\ref{fig:debug_overlay}).
\end{itemize}

\subsection{Rollout strategies}

I used multiple strategies to collect rollouts that differ in the garment-sampling logic, augmentation aggressiveness, and starting states (Table~\ref{tab:rollout_strategies}).

\begin{table*}[!htb]
\centering
\small
\begin{tabular}{@{}p{0.15\linewidth}p{0.75\linewidth}@{}}
\toprule
Strategy & What it does \\
\midrule
random & sample garments uniformly from the seen pool \\
full & all garments $\times$ N episodes \\
curriculum & prefer garments near a configurable target success rate --- the same difficulty-targeted selection idea as~\cite{sun2025difficulty} (which prefers training problems near a $\sim$50\% success rate), generalized to any target \\
success replay & restore a saved early-episode state of a successful episode and re-run (with heavier augmentations) \\
semi-success replay & restore a near-success state, replay open-loop to it, hand over to the live policy \\
hard mining & restore a saved failure state and try again \\
\bottomrule
\end{tabular}
\caption{Rollout-collection strategies.}
\label{tab:rollout_strategies}
\end{table*}

Only fresh, unbiased episodes --- \texttt{random}, \texttt{full}, and \texttt{curriculum} --- count toward the per-garment success-rate statistics; replay and hard-mining episodes would bias them (they are deliberately easier or harder than a fresh episode) but still contribute to training.

The replay strategies run on saved \textbf{physics states} --- particle positions/velocities plus joint states, snapshotted mid-episode and restorable later:

\begin{itemize}
    \item \textbf{Success states} are snapshotted at step 5 of episodes that end in success, saved with probability $1 - \mathrm{SR}$ (rare successes are always kept). Replays re-run them with extra augmentations and only successful replays are kept --- multiplying scarce successful data for hard garments.
    \item \textbf{Failure states} are snapshotted when the EMA-smoothed success prediction drops by more than 0.12 from its running max (when that max exceeded 0.25) --- i.e. at the moment the policy visibly ruined a promising episode. Hard mining restores these and rolls the dice again.
    \item \textbf{Semi-success states} come from failed episodes that reached the first checkpoint or sustained a high predicted success --- close-but-failed episdoes that can also be routed to the manual DAgger queue (\S3.4) instead of automatic retry.
\end{itemize}

\subsection{Environment augmentations}
\label{sec:env_aug}

I have implemented a visual/physical augmentation logic, applied at two scopes:

\begin{itemize}
    \item \textbf{Episode-level} (randomized at reset, saved with the physics state so replays can reproduce it): garment texture-pattern swap, LAB-space color remap, garment pose/scale/roughness perturbation, per-camera position/rotation/focal jitter, robot-base position jitter, table-texture transform, dome-light rotation. Physics-affecting components (scale, roughness, base jitter) are skipped when replaying saved states, which need the original dynamics.
    \item \textbf{Per-step} (re-randomized every few frames): garment color tint, arm color, dome-light intensity/color/temperature.
\end{itemize}

How aggressively to augment depends on who has to act on the augmented frames. In the first competition round, regular rollouts ran mostly in a \textit{light} regime --- pattern swap and color remap at $p{=}0.4$, $\pm$2 cm / $\pm$2$^\circ$ garment pose, $\pm$5\% scale, per-step color tint --- enough variety to regularize without degrading the live policy. Success replays are different: the policy's performance there doesn't matter, so the augmentations are much stronger for them. In the second (sim-to-real) round, collection moved to success replays only and went far more aggressive still --- swap/remap at $p{=}0.8$, camera pose and focal jitter calibrated against the real rig, arm-base shifts, table-texture transforms, per-step arm color and light temperature. That stack is what made sim rollouts usable as real-robot training data (sim-to-real section, \S9.8).

On the model side, light training-time image augmentations were always on, independent of the environment engine (\S2.6) --- and the sim-to-real phase made those much more aggressive too.

\subsection{Manual DAgger}

For failure modes the automated strategies couldn't fix, I built a DAgger-style~\cite{ross2011dagger} loop: load a saved failure or semi-success state into the sim, teleoperate for a few seconds to fix the garment, then hand control back to the policy --- if it finishes the fold, the whole episode is saved as a demonstration.

I made it as asynchronous as possible: hard states are saved during regular rollout collection, the human only spends a few seconds of the actual fix, and the corrected state is uploaded back to the Hub, where the next rollout cycle picks it up and treats it as a semi-success state. No human time is spent waiting for the policy to fail or to finish the task --- correction behavior can be collected non-stop, and multiple sims run in parallel so the operator never waits for loading.

I developed both a sim and a real-robot version of this tooling. The sim version turned out to be of limited use: teleoperating the robot through a sim interface is genuinely hard, and by the time I had it working the policy was simply better at folding than I was through teleop. The real-robot version, in contrast, became one of the most useful tools of the project --- see the sim-to-real section (\S9.10) for how real DAgger data is collected and weighted.

\section{Policy Architecture}
\label{sec:architecture}

I start from the architecture my team built for the BEHAVIOR-1K Challenge~\cite{behavior1k2025}, which itself extends $\pi_{0.5}$~\cite{black2025pi05}. The base stack: SigLIP-So400m/14 image encoder~\cite{zhai2023siglip} $\to$ Gemma-2B prefix transformer~\cite{gemma2024} (images + state + auxiliary tokens) $\to$ Gemma-300M action expert that generates a 30-step action chunk (1 s at 30 Hz, 12-dim joint deltas) via flow matching~\cite{lipman2023flow}, attending to the prefix KV cache. Three RGB cameras (top, left wrist, right wrist), all resized to 224$\times$224. The vision backbone is frozen. FAST action tokens~\cite{pertsch2025fast} in the prefix --- training-only auxiliary that shapes the VLM representation; absent at inference.

\textbf{Carried over from BEHAVIOR-1K} solution (not contributions of this work):

\begin{itemize}
    \item Text input is dropped entirely --- no tokenizer, no language tokens.
    \item Correlated flow-matching noise: the noise that seeds the denoising loop is drawn with the empirical action covariance (Cholesky factor from norm stats, shrinkage $\beta = 0.5$ toward identity).
    \item Correlation-aware soft inpainting at chunk boundaries: the tail of the previous chunk conditions the head of the next one through the same covariance structure, active only in the early, high-noise part of the denoising loop (while the flow time stays above a threshold tuned per garment type at inference, \S7.2), leaving the final low-noise steps free to self-correct.
    \item Cross-layer KV-cache mixing: before the action expert reads the prefix KV cache, each layer's K and V are replaced by a learned linear combination of all layers, letting the action expert choose which VLM depths to attend to.
    \item Multi-sample flow matching: 5 independent (noise, time) samples per batch item share one prefix forward pass.
\end{itemize}

The additions for LeHome are the token layout changes (\S4.1), the garment-type input token (\S4.2), advantage conditioning (\S4.3), multi-signal AdaRMS conditioning (\S4.4), XSA (\S4.5), and the smooth per-timestamp normalization of the action target (\S4.6). The auxiliary prediction heads that live on top of this layout are described in Section~\ref{sec:auxheads}.

I present the architecture as is. It was not properly ablated --- I was focused on the competition and result optimization rather than on proper research, and a real ablation study would multiply the budget (time and money). So I share all the details for reference without claiming which of them were actually critical.

\subsection{Model structure and attention layout}

\begin{figure*}[!htb]
  \centering
  \includegraphics[width=0.95\textwidth]{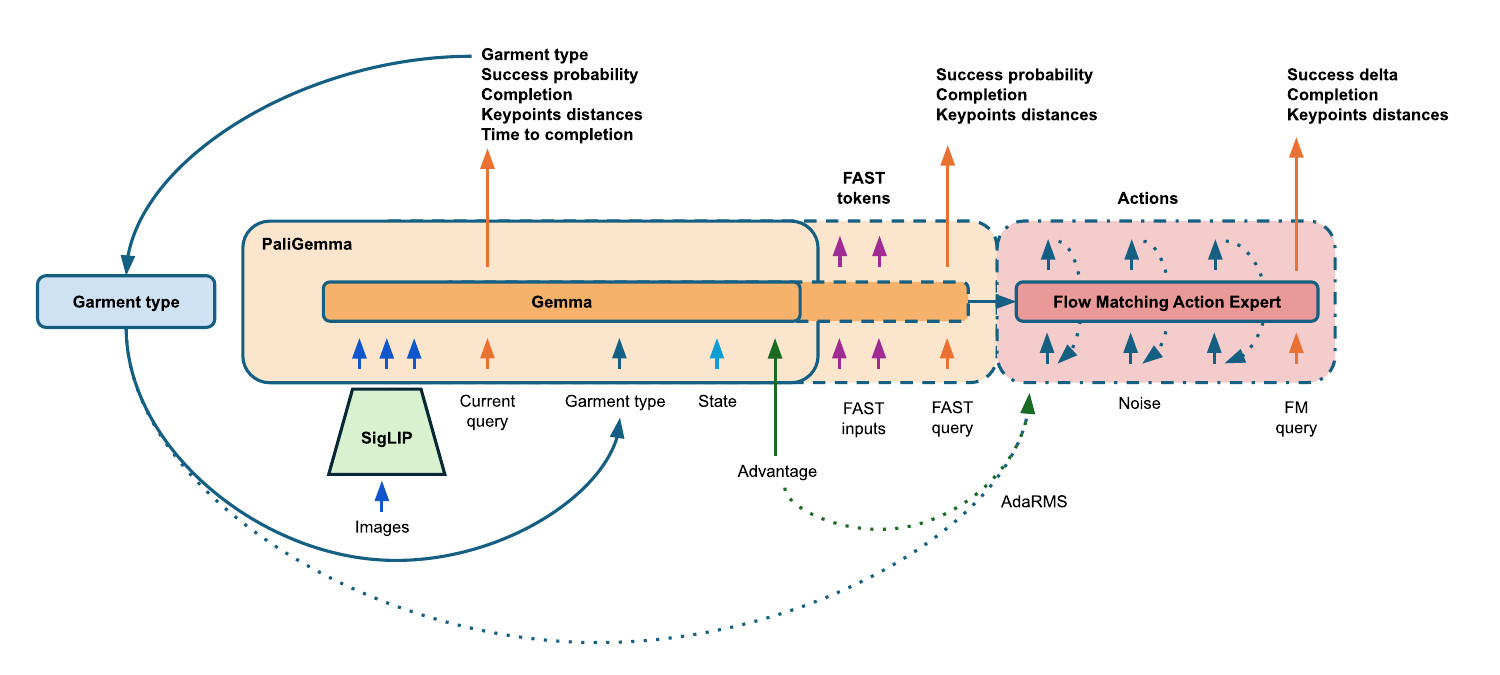}
  \caption{Policy architecture. A frozen SigLIP encoder feeds the Gemma-2B prefix (images, current query, state, garment type, advantage, FAST tokens, FAST query); the Gemma-300M flow-matching action expert reads the cleaned prefix and emits a 30-step action chunk. The prediction heads (orange) read from the prefix; the FM query at the tail of the action expert carries the action-conditional future predictions.}
  \label{fig:model_schema}
\end{figure*}

\begin{figure}[!htb]
  \centering
  \includegraphics[width=\linewidth]{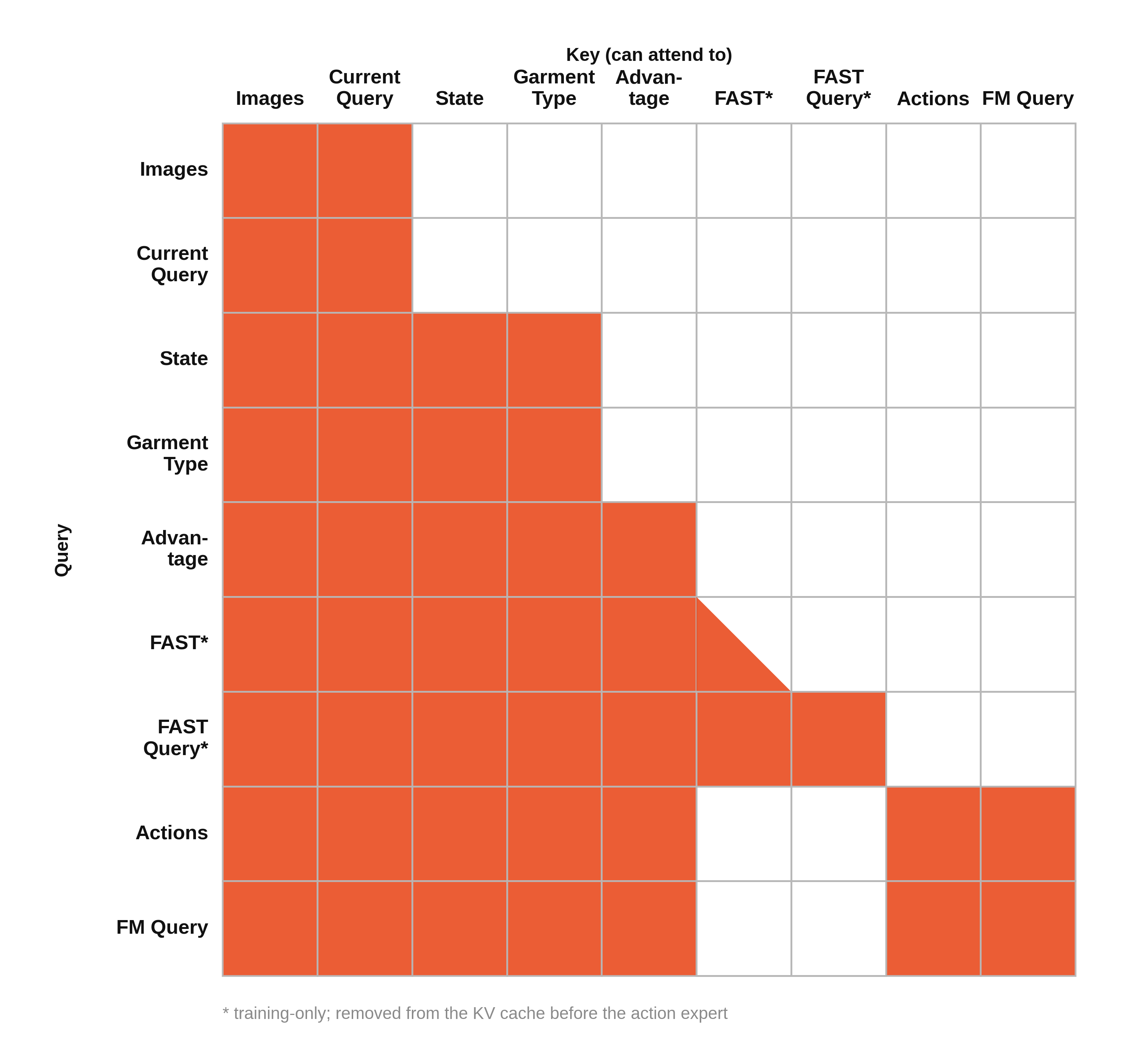}
  \caption{Attention mask.}
  \label{fig:attention_mask}
\end{figure}

Figure~\ref{fig:model_schema} gives the overall structure. The prefix combines several token groups under a hierarchical attention mask: each group sees itself and all groups before it, and earlier groups never see later ones (Figure~\ref{fig:attention_mask}).

\begin{itemize}
    \item \textbf{images} (3 cameras) together with the \textbf{current query} --- the single learned token all prediction heads read from (Section~\ref{sec:auxheads});
    \item \textbf{state} (12 joints, discretized into 256 bins and embedded through a dedicated 256-entry table) together with the \textbf{garment-type input token} (\S4.2);
    \item \textbf{advantage} (\S4.3), additionally maskable per-sample;
    \item \textbf{FAST tokens} (causal) and the \textbf{FAST query} --- training-only; both are removed from the KV cache before the action expert runs.
\end{itemize}

The action-expert suffix is 30 action tokens plus the \textbf{FM query}, fully bidirectional, attending to the whole (cleaned) prefix.

The key design choice is that the current query sits in the image group: state, garment type, and advantage are invisible to it, so every prediction head is forced to work from pixels alone --- the value heads cannot overfit to proprioceptive state, and the garment-type head cannot trivially copy the garment-type input token.

\subsection{Garment type input token}

A learned embedding (one of 4) placed alongside the state tokens. This token replaces the text prompt and tells the model which garment type is currently being folded.

The garment type is not provided as an input during evaluation, so I implemented a very basic System-2-like logic (similar to the stage-prediction logic in the BEHAVIOR-1K solution):

\begin{itemize}
    \item During training the ground-truth garment type is provided as the input.
    \item During evaluation the model predicts the garment type (via the auxiliary prediction head, Section~\ref{sec:auxheads}) at the very beginning of the episode and uses that prediction as the input for the rest of the episode.
\end{itemize}

I didn't ablate this, but I don't think this step is critical. At the late stages of training the garment-type prediction accuracy was above 99\% --- the model made mistakes mostly in very messy states where the folding had already failed. In most cases the model can identify the garment correctly from the images alone, and the extra garment-type input doesn't bring much additional information.

\subsection{Advantage conditioning}

During RL training every frame carries an advantage $A$ (Section~\ref{sec:reward}), and I condition the action expert on it --- the RECAP-style~\cite{recap2025} half of how advantage is consumed (the other half is sampling, Section~\ref{sec:training}).

\begin{itemize}
    \item A single learned ``advantage token'' sits in its own group: it sees images, state, and garment type, but is invisible to all of them.
    \item \textbf{Training}: if $A < 0$ the token is always masked out (neutral); if $A \geq 0$ it is masked stochastically, with $P(\text{neutral})$ ramping linearly from $0.5$ at $A{=}0$ down to $0.1$ at $A \geq 2$.
    \item The same gate drives the AdaRMS advantage vector (\S4.4), so the signal reaches the expert both through attention and through every layer norm.
\end{itemize}

\textbf{Inference}: the token is always active. CFG runs two action-expert passes over the shared prefix KV cache --- conditional and unconditional --- and combines the velocities as $\hat{v} = v_u + s\,(v_c - v_u)$ with a per-garment-type scale $s$ (\S7.3, \S7.7).

\subsection{AdaRMS multi-signal conditioning}

The action expert already modulates every layer's RMSNorm with a conditioning vector derived from the flow-matching time (AdaRMS, used in $\pi_{0.5}$). I add two more signals into the same vector:

{\footnotesize
\[ c \;=\; \underbrace{\mathrm{MLP}(\mathrm{posemb}(t))}_{\text{flow time (existing)}} \;+\; \underbrace{g[\text{garment\_type}]}_{\text{4 learned vectors}} \;+\; \underbrace{\mathbf{1}[\text{advantage active}]\cdot a}_{\text{1 learned vector}} \]
}

All new vectors are zero-initialized, so a checkpoint trained before this change resumes identically at step 0. The advantage term is gated by the same per-sample mask as the advantage token (\S4.3).

The motivation: in my early experiments the action predictions were not conditioned strongly enough by the garment type and advantage --- I observed the model performing movements for one garment while the input provided another (correct) one. A single token in the prefix is probably too weak a signal; AdaRMS is a much more direct one, modulating every layer of the action expert. I didn't run a proper ablation, but after adding the AdaRMS channel the wrong-garment behavior visually reduced.

\subsection{Exclusive Self-Attention (XSA)}

Standard attention lets every token trivially route its own value to the output by attending to itself. XSA~\cite{xsa2026} removes that path: after the attention step, the projection of the output onto the token's own value vector is subtracted,

\[ z_t \;=\; y_t - \frac{y_t \cdot v_t}{\lVert v_t \rVert^2 + \varepsilon}\, v_t, \]

where $y_t$ is the standard attention output and $v_t$ the token's own value. It is applied to both the VLM and the action expert.

It is a very new work - just a few months old. The initial results show XSA improves training and validation loss at negligible compute overhead, so I tested it. Adding it to a pretrained model initially breaks the loss, but it recovered surprisingly fast, so I kept it. No real ablation was done --- I rely on ``recent fashion'' here, and since I post-train the model heavily anyway and don't care about catastrophic forgetting, I assume it should not hurt.

\subsection{Action normalization}

The chunk is predicted in a normalized space and the flow-matching loss is computed there, so normalization shapes the target. Actions are deltas from the current state, a delta's spread grows with the horizon --- roughly a random walk (std $\propto \sqrt{t}$), though the real shape varies per dimension and carries an extra first-step shift from the gap between the last executed action and the measured state. I therefore normalize \textbf{per timestep} --- each horizon position gets its own mean and std --- rather than with one global scale, which would over-shrink the near steps (the ones we actually execute and re-plan from~\cite{lbm2025}) and let the high-variance tail dominate the loss. The empirical per-timestep std can be a bit noisy across the horizon, so instead of the raw statistics I scale by a smooth per-dimension fit,

\[ \sigma_d(t) \;=\; a_d + s_d\,\sqrt{t + e_d}\,, \]

(with a linear fit for the mean), which keeps the per-timestep magnitude profile while leaving a smooth trajectory smooth.

\section{Auxiliary Prediction Heads}
\label{sec:auxheads}

I use two groups of auxiliary prediction heads: current-frame heads (\S5.1) that include the success and completion predictions used as a value function, and future-prediction heads (\S5.2) that act as a Q-function and a cheap quasi-world model, predicting what the scene will look like 30 frames ahead.

All current-frame prediction heads are linear probes off a single learned \textbf{current query} token (1$\times$2048) placed in the image attention group (\S4.1): they see the images and nothing else, and they all share one VLM forward pass.

All future-prediction heads are duplicated --- one is placed after the FAST tokens in the prefix, the other at the tail of the action-expert suffix --- and both can attend to the corresponding action-representation tokens.

Keeping the value / Q function inside the policy model also simplifies the training pipeline --- there is only one model to train, serve, and version. And conceptually, predicting success probability or the future state requires many of the same primitives as deciding on the best action chunk, so these ``models'' share a lot of logic anyway: putting them in one network saves compute, provides useful auxiliary signal to the shared representation, and allows some positive cross-learning.

All head gradients flow into the VLM backbone (no stop-gradient); small loss weights keep them from dominating.

\subsection{Current-frame heads}

The current-frame heads are summarized in Table~\ref{tab:current_heads}.

\begin{table*}[!htb]
\centering
\small
\begin{tabular}{@{}p{0.18\linewidth}p{0.16\linewidth}p{0.3\linewidth}p{0.22\linewidth}@{}}
\toprule
Head & Output & Target & Trained on \\
\midrule
success & sigmoid & P(episode succeeds) & RL rollouts \\
checkpoint & sigmoid & P(reach mid-episode checkpoint) & RL rollouts \\
garment type & softmax(4) & garment class & all data \\
completion & sigmoid & $t/T$ & successful episodes only \\
TTC & sigmoid & time-to-completion (TD) & all data \\
keypoint distances & raw linear (21) & per-condition distance ratios & RL rollouts \\
\bottomrule
\end{tabular}
\caption{Current-frame auxiliary heads.}
\label{tab:current_heads}
\end{table*}

\textbf{Success head} --- the main one: it is the value function of the whole RL setup. It predicts the probability that the current episode ends in success, and drives advantage computation (Section~\ref{sec:reward}), and live failure detection during rollouts. Trained with BCE on the binary episode outcome.

\textbf{Completion head} --- the source of the potential-based progress shaping (Section~\ref{sec:reward}), the second per-frame signal alongside the success value: predicts the fraction of the episode completed, $t/T$, trained on successful episodes only. Unlike the success head it is policy-stable --- task progress looks almost the same no matter which policy produced it.

\textbf{Garment-type head} --- 4-class classifier that feeds the inference-time garment-type bootstrap (\S4.2).

\textbf{Checkpoint head} --- predicts the probability of reaching the mid-episode checkpoint (\S6.1). Legacy: the predictions are not used in the final solution.

\textbf{Keypoint distance head.} The challenge's success checker compares keypoint-pair distances against per-garment thresholds (\S6.1). I predict exactly those ratios $d^{(i)} = \mathrm{dist}_i/\mathrm{threshold}_i$ from the image: 21 outputs organized as per-garment-type slices (top\_long [0:5], top\_short [5:10], pant\_long [10:17], pant\_short [17:21]); only the slice matching the current garment is active in the loss, the rest are NaN-masked.

The motivation came from world models: I wanted to encode the current state and predict the future state conditioned on actions. But a full world model is very expensive, and most of the state is not worth predicting --- the robot state is trivial, and only the garment state carries value. Instead of pixel-level prediction or encoding/decoding through a latent space, I predict the few numbers that matter most about the garment: the same keypoint distances that define success. It is a very cheap world-model substitute --- with the caveat that it is generally only available in simulation, since the targets require privileged data.

\textbf{Time-to-completion head.} The target is a 30-step TD bootstrap: $\tau_t = \hat{\tau}_{t+30} - 30/600$ on success, $\hat{\tau}_{t+30}$ on failure. This is a legacy head left from past experiments: it is aligned with the $\pi^{*}_{0.6}$ approach and effectively combines success probability and completion percentage in one scalar (Section~\ref{sec:reward}). I wanted to test it as an alternative value function, but the results were inconclusive and I ran out of time for further investigation --- so it stays as a training target but is not used anywhere downstream.

\subsection{Future-prediction heads}

I also predict the same quantities 30 frames ahead, from two extra query tokens:

\textbf{FAST query (training-only).} Appended after the FAST block in the prefix, in its own attention group (it sees everything including FAST; nothing sees it). Predicts success, completion, and keypoint distances at $t{+}30$. Since FAST is absent at inference, this head exists purely to push future-awareness into the VLM representation during training; it is stripped from the KV cache before the action expert runs.

\textbf{FM query.} Appended at the tail of the action-expert suffix, bidirectional with the 30 action tokens --- so it reads the \textit{denoised actions} at every flow step and its predictions are conditioned on what the policy is about to do.

It predicts the future keypoint distances --- this is where the ``world model'' actually sits. It is not a real world model, but it works as a very cheap analogue: many practical applications extract some kind of reward from world-model predictions anyway, so why not predict that reward-relevant state directly?

It also predicts a success \textbf{residual}:

\[ \Delta_{\text{success}} \;=\; y \;-\; \mathrm{sg}\!\left(\hat{P}_{\text{success}}\right), \]

the true outcome minus the (stop-gradient) current success-head estimate: ``given these specific actions, how much better or worse than the image-only baseline will this end?'' Raw linear output, MSE loss. The stop-gradient keeps this Q/Advantage-like head from dragging the V-like success head.

Flow-time weighting: the loss is scaled by $(1-t)$ and samples with $t > 0.5$ are excluded entirely --- a mostly-noise action chunk carries no usable signal about the future, and clean-action samples should dominate.

\subsection{Loss weights and training details}

The flow-matching action loss has weight 1.0. The auxiliary-loss weights in the final simulation configuration are shown in Table~\ref{tab:loss_weights}.

\begin{table}[!htb]
\centering
\small
\begin{tabular}{@{}lr lr@{}}
\toprule
 & weight & & weight \\
\midrule
success & 0.05 & keypoint\_distance & 0.02 \\
ttc & 0.05 & wm\_fast ($\times$3) & 0.01 \\
completion & 0.02 & wm\_flow success & 0.1 \\
garment\_type & 0.02 & wm\_flow completion & 0.05 \\
checkpoint & 0.001 & wm\_flow keypoint & 0.02 \\
FAST & 0.01 & & \\
\bottomrule
\end{tabular}
\caption{Auxiliary-loss weights (final simulation configuration).}
\label{tab:loss_weights}
\end{table}

Extra details:
\begin{itemize}
    \item \textbf{Success tail boost.} The last 20 frames of successful episodes get a 20$\times$ weight on the success BCE. When the policy fails very close to success it stays in that state for a long time and generates many ``almost success but fail'' training frames, while successful episodes stop immediately. This drives the success prediction to drop sharply just before an actual success; the boost rebalances the frequency of these visually identical ``almost done'' states (the training-time complement of the offline value tail correction, \S6.2).
    \item \textbf{Regularization.} The aux heads tend to massively overfit on the training data, so an extra weight decay of 0.001 is applied to their kernels (on top of the near-zero base decay).
    \item \textbf{Label smoothing.} Success targets are smoothed toward the per-garment average success rate: $y' = y(1-\alpha) + \bar{p}_g\,\alpha$ with $\alpha = 0.05$.
\end{itemize}

\subsection{Where the predictions are used}

\begin{itemize}
    \item \textbf{success + completion} $\to$ advantage computation (Section~\ref{sec:reward}) and live failure detection during rollouts (Section~\ref{sec:data}).
    \item \textbf{garment type} $\to$ the inference-time garment-type bootstrap (\S4.2).
    \item \textbf{$\Delta_{\text{success}}$} $\to$ best-of-N action selection at inference: several candidate chunks are denoised in parallel from the shared prefix cache, and the one with the best predicted $\Delta_{\text{success}}$ is executed (details and caveats in Section~\ref{sec:inference}).
\end{itemize}

\section{Reward Design and Advantage Computation}
\label{sec:reward}

Reward engineering is one of the most important steps in RL training. While advantage estimation given a reward is usually theoretically grounded, the choice of reward itself is partly an ``art'' built on heuristics and experiments. This section explains how I arrived at the final advantage function, step by step, including the subjective choices.

The competition metric is a binary episode success --- too sparse for efficient RL: correct actions early in the episode get almost no signal, and an easy success is indistinguishable from a hard one. Common densification approaches include learned value/Q functions, hand-crafted partial-progress rewards, completion or time-to-go prediction, and group-relative baselines (GRPO~\cite{shao2024grpo}). My final setup combines elements of all of them.

\textit{Note: the simulator also provides a shaped reward reflecting the ``niceness'' of the final garment configuration; I did not use it.}

\subsection{Dense reward from the success checker}

The challenge defines success via per-garment keypoint-distance conditions (5 for tops, 4 for short pants, 4 for long pants --- which I expanded to 7): each condition compares a distance ratio $d^{(i)} = \mathrm{dist}_i/\mathrm{threshold}_i$ against 1 (proximity conditions need $d^{(i)} \le 1$, spread conditions $d^{(i)} \ge 1$). I reuse exactly these conditions to build intermediate checkpoints, so no new keypoints or success definitions are needed. Both shirt types and long pants get an extra intermediate checkpoint; short pants are the easiest type, so no extra checkpoint.

Reaching the intermediate fold checkpoint is worth 0.5 reward; the full success checkpoint brings the cumulative reward to 1.0.

\textbf{Gradual first checkpoint (applied to tops only).} To densify further, the first $0.5$ is not granted as a single spike but allocated in proportion to the reduction of the primary proximity distance. With $d_t$ the primary distance ratio, $m_t = \min_{\tau \le t} d_\tau$ its running minimum, and $t_1$ the frame where the checkpoint is first reached, the cumulative allocated reward is

\[\resizebox{\columnwidth}{!}{$\displaystyle R^{\mathrm{cp1}}_t \;=\; 0.5\,\mathrm{clip}\!\left(\frac{d_0 - m_t}{d_0 - d_{t_1}},\,0,\,1\right), \qquad r_t \mathrel{+}= R^{\mathrm{cp1}}_t - R^{\mathrm{cp1}}_{t-1}.$}\]

\begin{figure*}[!htb]
  \centering
  \includegraphics[width=0.9\textwidth]{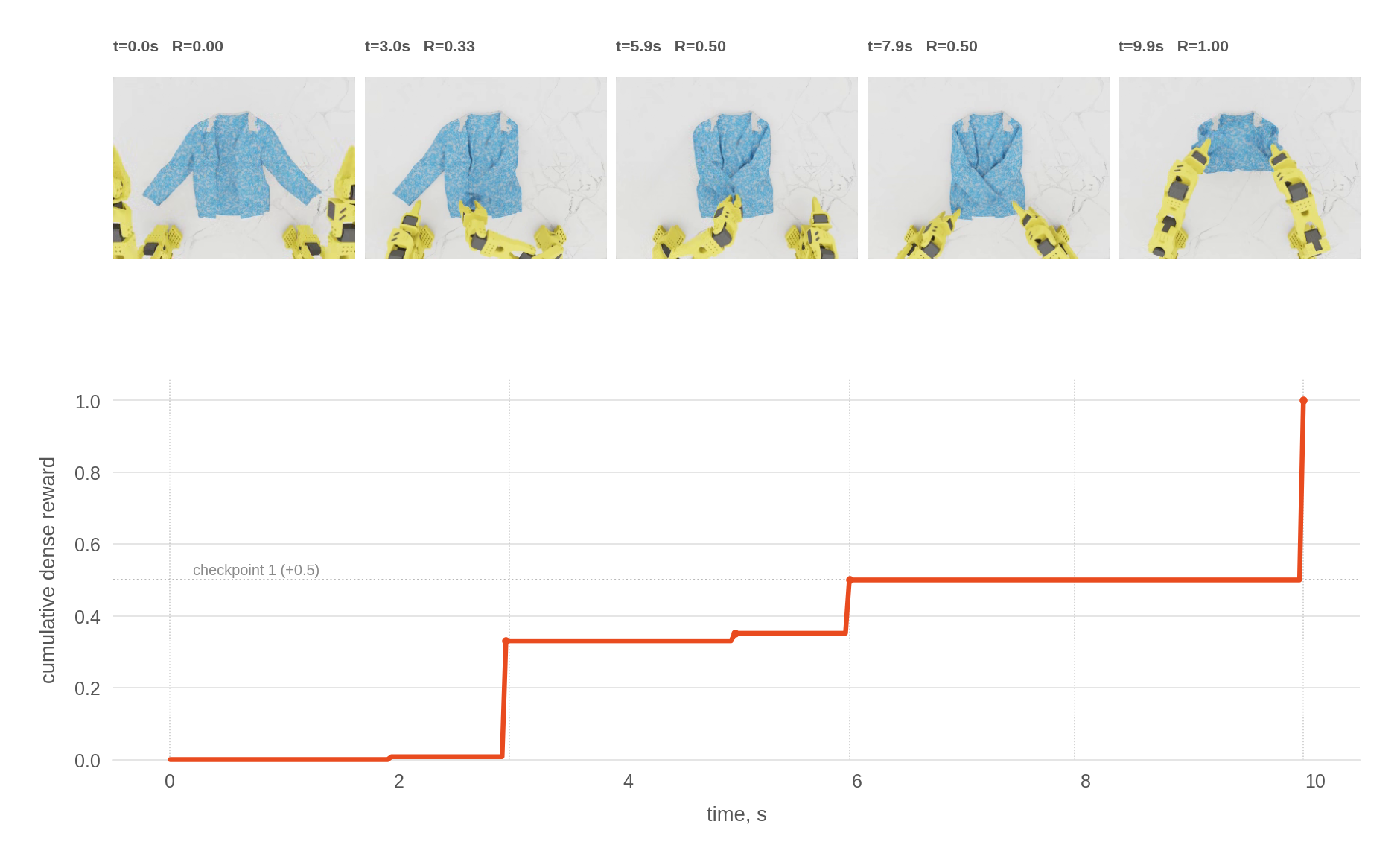}
  \caption{Dense reward on a successful long-sleeve-top episode: overhead frames at five milestones (top) and the cumulative reward (bottom). The first $0.5$ is allocated gradually as the primary proximity distance closes --- the gradual first checkpoint --- and reaching full success brings the total to $1.0$.}
  \label{fig:dense_reward}
\end{figure*}

E.g. if folding the first sleeve already closes 60\% of the distance gap, it earns $0.3$ of the reward (Figure~\ref{fig:dense_reward}).

\textbf{Failure withdrawal.} If the episode fails, all accumulated reward is withdrawn so the total return is always binary: $\sum_t r_t =1[\text{success}]$. Rationale: reaching a checkpoint can be misleading --- the checkpoint conditions may hold while the rest of the garment is in a state that makes success unlikely. To avoid one sharp negative spike at the end, the withdrawal is spread uniformly from the last frame $t_p$ at which the cumulative reward reached its maximum:

\[ r_t \mathrel{-}= \frac{\sum_\tau r_\tau}{T - t_p - 1}, \qquad t_p < t < T. \]

Intermediate checkpoints therefore provide temporal credit \textit{within} an episode, while the episode-level return stays aligned with the true objective.

\subsection{Success probability as a value function}
\label{sec:value}

Because the total return equals the success indicator, the value function is simply

\[ V(s_t) \;=\; \mathbb{E}[\text{reward remaining}] \;=\; P(\text{success}\mid s_t) - R^{\mathrm{cum}}_t, \]

so approximating it only requires a success-probability predictor. A separate value network is the standard choice, but it complicates the pipeline and wastes compute; instead I train $\hat{P}_t$ as an auxiliary head of the VLA itself (Section~\ref{sec:auxheads}): a shared query token that attends only to image tokens (preventing state-shortcut overfitting).

The classical use of a value baseline is to subtract it from the return to reduce variance~\cite{schulman2015gae}. Two problems with applying $\text{return} - V(s_t)$ literally here:

\begin{enumerate}
    \item \textbf{Checkpoint rewards cancel.} With $V_t = P(\text{success}\mid s_t) - R^{\mathrm{cum}}_t$ and $R^{\mathrm{cum}}_{t+1} = R^{\mathrm{cum}}_t + r_t$, the TD residual collapses to
    \[\resizebox{\linewidth}{!}{$\displaystyle \delta_t = r_t + V_{t+1} - V_t = P(\text{success}\mid s_{t+1}) - P(\text{success}\mid s_t).$}\]
    The reward enters both the remaining return and $R^{\mathrm{cum}}_t$ and the two offset, so the dense reward I just built disappears from the advantage entirely.
    \item \textbf{The variance argument assumes a perfect predictor.} What is actually subtracted is a model prediction $\hat{V}$, not $V$. With an imperfect $\hat{V}$, full subtraction is no longer the variance-minimizing choice.
\end{enumerate}

\textbf{A dampened baseline.} Treat the prediction as a noisy control variate and subtract only a fraction $\alpha$ of it,
\[ \hat{A}_t = G_t - \alpha\,\hat{V}_t, \qquad G_t = 1[\text{success}] - R^{\mathrm{cum}}_t, \]
where $G_t$ is the observed return-to-go. The variance is a quadratic in $\alpha$,
\[\resizebox{\columnwidth}{!}{$\displaystyle \operatorname{Var}\big(G - \alpha\hat{V}\big) = \operatorname{Var}(G) - 2\alpha\operatorname{Cov}(G,\hat{V}) + \alpha^{2}\operatorname{Var}(\hat{V}),$}\]
minimized at
\[ \alpha^{*} = \frac{\operatorname{Cov}(G,\hat{V})}{\operatorname{Var}(\hat{V})} = \frac{\operatorname{Cov}(V,\hat{V})}{\operatorname{Var}(\hat{V})} = \rho(V,\hat{V})\,\frac{\sigma(V)}{\sigma(\hat{V})}. \]
The second equality holds because the outcome noise $G - V$ has zero mean given the state and is therefore uncorrelated with any function of the state, $\hat{V}$ included; the first form contains only observable quantities. To see what $\alpha^*$ looks like, model the head as the true value plus an independent error, $\hat{V} = V + \varepsilon$ with $\varepsilon$ independent of $(V, G)$ and $\operatorname{Var}(\varepsilon) = \sigma_\varepsilon^{2}$. Then $\operatorname{Cov}(G,\hat{V}) = \sigma_V^{2}$, $\operatorname{Var}(\hat{V}) = \sigma_V^{2} + \sigma_\varepsilon^{2}$, and
\[ \alpha^{*} = \frac{\sigma_V^{2}}{\sigma_V^{2} + \sigma_\varepsilon^{2}} = \rho^{2}(V,\hat{V}) < 1. \]

A perfect predictor gives $\alpha^{*} = 1$ and recovers full subtraction - equivalent to the standard advantage estimation; a useless one gives $\alpha^{*} = 0$ and falls back to the raw return.

\textbf{Both problems are one problem.} Carrying the fraction into the TD residual (with $\gamma = 1$ for clarity; the exact $\gamma$-dependent form is in Section~\ref{sec:gae}) gives
\[ \delta_t = (1-\alpha)\,r_t + \alpha\,\big(\hat{P}_{t+1} - \hat{P}_t\big), \]
so the dense reward and the change in predicted success probability are mixed in proportion to how much the predictor can be trusted. $\alpha = 0$ leaves the pure checkpoint reward. $\alpha = 1$ is exactly the cancellation of problem 1 --- which is now the correct behaviour rather than a bug: if the success probability were predicted perfectly, intermediate rewards would carry no extra information, and the change in $P(\text{success})$ would be the whole signal. Problem 1 was an artifact of assuming a perfect predictor. 
$G - \alpha\hat{V}$ with $\alpha < 1$ is no longer an unbiased estimate of the advantage; but I keep the name because it plays the advantage's role downstream.

The idea was inspired by online experimentation variance reduction approach. CUPED~\cite{deng2013cuped} reduces A/B-test variance by subtracting from each unit's metric $Y$ a covariate $X$ that predicts it, scaled by a coefficient $\theta$:
\[ Y_{\mathrm{adj}} = Y - \theta\,\big(X - \mathbb{E}[X]\big), \qquad \theta^{*} = \frac{\operatorname{Cov}(Y,X)}{\operatorname{Var}(X)}, \]
that is the problem mathematically equivalent to ours. The original paper uses the pre-experiment value of the metric as $X$; the form used in industry today usually uses the best available prediction of the outcome, which is exactly the value-prediction setup.

The idea itself is the textbook control-variate / regression-adjustment result and is broadly known, but it is not how mainstream advantage estimation is usually done in RL papers.

\textbf{What I actually used.} Since the estimation of $\alpha^{*}$ needs only returns and predictions, it is directly estimable from rollouts, including separately per garment type, provided the predictions are made on data the head has not trained on (otherwise the head has partly memorized the outcome noise and the naive estimate is biased toward $1$). My pipeline records $\hat{P}_t$ at collection time (Section~\ref{sec:stale}), which is exactly such an on-policy, unseen-state measurement. During the competition I nevertheless made a logical mistake here (carried into v1 of this report): I believed $\alpha^{*}$ needed the true $V$ and was therefore not computable, and used a fixed coefficient chosen by judgement instead:
\[ \hat{V}_t = \alpha_s\big(P(\text{success}\mid s_t) - R^{\mathrm{cum}}_t\big), \qquad \alpha_s = 0.5. \]
Post-factum estimates of $\alpha^{*}$ on the historical rollouts vary between roughly $0.4$ and $0.8$ across garment types and training stages, with $0.5$ close to the median, so the mistake was not critical. A dynamic, per-garment $\alpha^{*}$ re-estimated from each iteration's fresh rollouts is what I would recommend now.

Two corrections make the raw head predictions more usable:

\begin{itemize}
    \item \textbf{EMA smoothing.} Raw per-chunk predictions are noisy; all consumers use $\bar{S}_t = \mathrm{EMA}(\hat{P}_t)$ with $\alpha_{\mathrm{EMA}} = 0.2$.
    \item \textbf{Value tail correction.} The head systematically under-predicts near successful endings: ``almost done'' frames overwhelmingly come from \textit{failed} episodes (a success terminates immediately; a near-miss stays in that state for many frames). Offline --- where the outcome $y \in \{0,1\}$ is known --- I replace the last $K{=}30$ frames of $\bar{S}$ with a linear interpolation toward $y$:
    \[\resizebox{\linewidth}{!}{$\displaystyle \bar{S}_{T-K+i} \;=\; \bar{S}_{T-K-1} + \left(y - \bar{S}_{T-K-1}\right)\frac{i}{K}, \qquad i = 1,\dots,K.$}\]
    The same bias is also attacked at training time: the last 20 frames of successful episodes get a $20\times$ weight on the success BCE loss.
\end{itemize}

\subsection{Completion prediction}

The success head alone has three weaknesses: (1) it is noisy and overfits to particular states; (2) it gives no signal on high-success garments, where it quickly saturates above 90\%; (3) it drifts during training since the same network predicts the actions (by design, but destabilizing). I therefore add a \textbf{completion head} --- trained with MSE on the target $t/T$ using successful episodes only --- and use it as a second progress signal. Completion is far more stable than success probability: it barely changes as the policy evolves, and it keeps providing signal even when $P(\text{success}) \approx 1$.

\subsection{GAE over both heads}
\label{sec:gae}

Both signals are aggregated with GAE~\cite{schulman2015gae} ($\gamma = 0.999$, $\lambda = 0.99$), computed offline in a separate pass before each training iteration.

\textbf{Success GAE} --- the exact TD residual of the $\alpha_s$-dampened value baseline $\hat{V}_t = \alpha_s\big(\bar{S}_t - R^{\mathrm{cum}}_t\big)$ (Section~\ref{sec:value}), with the terminal value pinned to the true outcome ($\bar{S}_T = y$):

\[\resizebox{\columnwidth}{!}{$\displaystyle \delta^{s}_t = r_t + \gamma\,\hat{V}_{t+1} - \hat{V}_t, \qquad A^{s}_t = \delta^{s}_t + \gamma\lambda\,A^{s}_{t+1}, \qquad \alpha_s = 0.5.$}\]

Expanding $\hat{V}$ (using $R^{\mathrm{cum}}_{t+1} = R^{\mathrm{cum}}_t + r_t$) and collecting the $r_t$ terms shows the reward is damped by the dampened return-subtraction:

\[\resizebox{\columnwidth}{!}{$\displaystyle \delta^{s}_t = \big(1-\alpha_s\gamma\big)\,r_t + \alpha_s\big(\gamma\,\bar{S}_{t+1} - \bar{S}_t\big) + \alpha_s(1-\gamma)\,R^{\mathrm{cum}}_t.$}\]

With $\gamma \approx 1$ the $\alpha_s(1-\gamma)R^{\mathrm{cum}}_t$ term vanishes, leaving a simple damping of the reward by $(1-\alpha_s\gamma)$:

\[ \delta^{s}_t \approx \big(1-\alpha_s\gamma\big)\,r_t + \alpha_s\big(\gamma\,\bar{S}_{t+1} - \bar{S}_t\big). \]

\textbf{Completion shaping} --- a \emph{potential-based shaping} term over the EMA-smoothed completion prediction $\bar{C}_t$ with potential $\Phi_t = \alpha_c\bar{C}_t$ and terminal $\bar{C}_T = y$:

\[ \delta^{c}_t = \gamma\,\Phi_{t+1} - \Phi_t = \alpha_c\left(\gamma\,\bar{C}_{t+1} - \bar{C}_t\right), \qquad \alpha_c = 0.5, \]
\[ \Phi^{c}_t = \delta^{c}_t + \gamma\lambda\,\Phi^{c}_{t+1}. \]

\subsection{Stale rollouts: segment baselines and blending}
\label{sec:stale}

The GAE above cannot be applied uniformly in a continuous, off-policy data-collection loop. $P(\text{success})$ is policy-dependent, and the predicting model both evolves between iterations and overfits to data it has already trained on (predictions on past rollouts become extreme and useless). My solution:

\begin{enumerate}
    \item \textbf{Predict at collection time.} Success/completion predictions are recorded during the rollout --- on-policy and on unseen states --- and never re-predicted later.
    \item \textbf{Decay old data.} Each rollout dataset's sampling share decays by $0.98$ per training iteration; BC and DAgger datasets keep fixed shares.
    \item \textbf{Blend toward an objective baseline.} As predictions go stale, the advantage shifts from GAE to a GRPO-style~\cite{shao2024grpo} relative-success signal that depends only on outcomes, not on the policy's predictions:
\end{enumerate}

\[ \tilde{A}_t = w\,A^{s}_t + (1 - w)\,A^{\mathrm{seg}}_t + \Phi^{c}_t, \]
\[ w = \min(\text{sampling share},\,1). \]

The completion shaping $\Phi^{c}$ stays at full strength for all data --- the completion head is policy-stable, so it remains valid for old rollouts. (It is added as a potential-based shaping term, not a value baseline, so it never needs the $R^{\mathrm{cum}}$ correction the success channel does.)

\textbf{Segment component $A^{\mathrm{seg}}$.} Episodes are split at the first checkpoint; each segment's return is compared against a per-garment empirical baseline ($p_{\mathrm{cp}}$ is the garment's checkpoint rate, SR its success rate), and scaled by $G(n)/n$ where $G(n) = \frac{1 - (\gamma\lambda)^n}{1 - \gamma\lambda}$ matches per-step magnitudes to GAE:

\[\resizebox{\columnwidth}{!}{$\displaystyle A^{\mathrm{seg}}_t = \begin{cases} \left(R_1 - \tfrac{1}{2}\,p_{\mathrm{cp}}\right) G(n_1)/n_1 & t \le t_1 \\[2pt] \left(R_2 - \left(\mathrm{SR}/p_{\mathrm{cp}} - \tfrac{1}{2}\right)\right) G(n_2)/n_2 & t > t_1 \end{cases}$}\]

and $(R - \mathrm{SR})\,G(T)/T$ for episodes that never reach a checkpoint (baseline fallback: garment mean $\to$ type mean $\to$ 0.5). Key property: reaching the checkpoint and then failing yields \textit{positive} advantage before the checkpoint and \textit{negative} after. 

\textbf{Normalization.} The blended $\tilde{A}_t$ is divided by a single global std $\sigma$ computed over unbiased rollouts only then clipped to $[-2, 2]$.

\subsection{Precision boost}

Binary success treats a barely-passing fold and a tight fold identically. After advantages are computed, the top 20\% of successful episodes per garment --- ranked by how tightly they satisfied the success conditions --- get a fixed bonus $\Delta A = 0.3$ on every frame. The tightness score is the worst (minimum) per-condition margin on the final frame's distance ratios, where each margin measures how far inside its threshold a condition sits ($1 - d^{(i)}$ for proximity conditions, $d^{(i)} - 1$ for spread conditions). Garments with fewer than 5 successes in the window are skipped. This biases training toward high-quality folds rather than marginal ones.

\subsection{How the advantage is used}
\label{sec:advantage_use}

\begin{figure}[!htb]
  \centering
  \includegraphics[width=\linewidth]{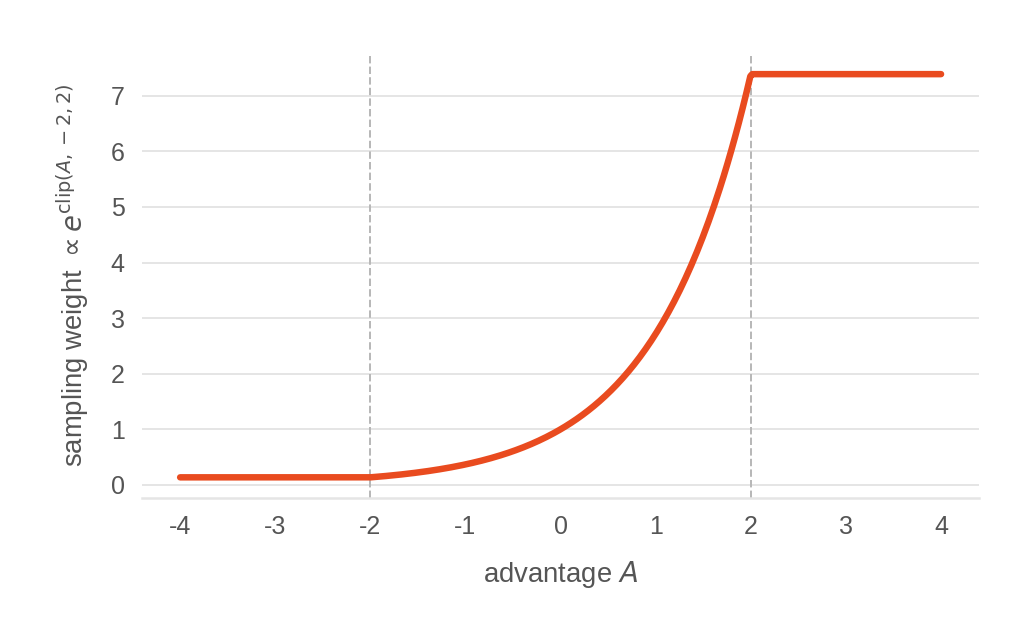}
  \caption{AWR sampling weight as a function of advantage, $P \propto e^{\mathrm{clip}(A,-2,2)}$.}
  \label{fig:sampling_weight}
\end{figure}

The advantage drives training in two ways (details in Section~\ref{sec:training}): AWR-style~\cite{peng2019awr} prioritized sampling, $P(\text{frame } i) \propto e^{\mathrm{clip}(A_i,\,-2,\,2)}$ (Figure~\ref{fig:sampling_weight}), and RECAP-style~\cite{recap2025} advantage conditioning of the action expert, which enables classifier-free guidance at inference.

\subsection{Intuitive summary}

The final formula is over-engineered and could probably be simplified without losing much. The short version --- an action's advantage is high when:

\begin{itemize}
    \item it makes objective task progress, measured by the same keypoint distances that define success;
    \item the predicted probability of eventual success increases;
    \item conditioned on eventual success, predicted completion increases.
\end{itemize}

The staler a rollout, the more its advantage degrades gracefully toward a sparse, outcome-only relative-success signal. Everything is aggregated with a GAE-like backward pass with $\gamma\lambda \approx 0.989$.

\section{Inference-Time Optimization}
\label{sec:inference}

The same model checkpoint can achieve different success rates depending on \textit{how} it is run: how much of each predicted chunk is actually executed, how strongly the advantage conditioning is amplified, how many candidate chunks are drawn and re-ranked. None of these touch the weights --- they are all decided at deployment, and they make a large difference. This section covers every knob applied at inference and how I tuned them per garment type.

The policy server is \textbf{stateless} --- each inference request is independent and carries no memory between calls (\S2.1, \S3.1). Everything that \textit{looks} stateful at inference --- the rolling chunk cache, the inpaint anchor, the latched garment type --- lives on the client side (the sim proxy or the real-robot runner). The server just maps an observation to a chunk.

\subsection{The inference contract: chunk execution and denoising}

The policy always predicts a fixed chunk of $H = 30$ actions, produced by a short flow-matching denoising loop ($S = 10$ Euler steps, integrating from noise at $t{=}1$ to clean actions at $t{=}0$). How that chunk is \textit{consumed} is governed by three quantities, all carried over unchanged from my team's earlier BEHAVIOR-1K solution~\cite{behavior1k2025}:

\begin{itemize}
    \item \textbf{Execution length} $n_e$ --- how many of the $H$ actions are actually sent to the robot before re-planning. A smaller $n_e$ means more frequent re-planning (more reactive, more compute); a larger $n_e$ is more open-loop.
    \item \textbf{Playback stretch} $k$ --- the executed actions are time-rescaled to $\lfloor k\,n_e\rfloor$ control steps. With $k>1$ the same motion is stretched over more steps, so the arm moves more slowly.
    \item \textbf{Anchor length} $n_a$ --- how many actions past the executed slice are retained to seed the next chunk (\S7.2), subject to $n_e + n_a \le H$.
\end{itemize}

\subsection{Soft inpainting between chunks}

Re-planning from scratch at every window would make the trajectory jump at each chunk boundary. Instead, the trailing $n_a$ actions of the current chunk become a soft anchor for the next one. While the next chunk denoises, the anchored action dimensions are gently pulled toward those anchor values along the flow path, and that pull is propagated to the remaining dimensions through the action covariance, so the \textit{whole} chunk stays consistent --- not just the overlapping part. The anchoring is active only in the first, high-noise part of the denoising loop --- while the flow time stays above a threshold $t_\text{ip}$ (recall $t{=}1$ is pure noise, $t{=}0$ the clean action) --- and switches off for the final low-noise steps. So the anchor sets the overall direction early, but the chunk is always left free to sharpen and self-correct at the end of the flow rather than being pinned to the anchor. The detailed mechanism is inherited from the BEHAVIOR-1K solution~\cite{behavior1k2025}.

I expect several benefits from it:

\begin{itemize}
    \item \textbf{Mode stickiness} --- the policy stays in the behavioral mode it already committed to, instead of jumping to a different one each chunk.
    \item \textbf{Smooth trajectories} --- which improves stability and also lets the model be trained at the frame level rather than the chunk level: any 30-step window of data is itself a smooth chunk.
    \item \textbf{Throughput} --- real-time chunking (RTC)~\cite{rtc2025} and related approaches show that this kind of overlap improves the effective throughput of the model.
\end{itemize}

\subsection{Classifier-free guidance on the advantage}

At inference I want the policy's best behavior, not its average. Classifier-free guidance~\cite{ho2022cfg} amplifies that conditioning. Every denoising step runs the action expert twice:

\begin{itemize}
    \item \textbf{conditional} --- the advantage signal present in both the prefix and the action-expert conditioning;
    \item \textbf{unconditional} --- the advantage signal removed from both,
\end{itemize}

and the two velocity fields are extrapolated:

\[ \hat{v} = v_\text{uncond} + \alpha\,\big(v_\text{cond} - v_\text{uncond}\big). \]

The two passes share the prefix computation (the expensive vision-language forward runs once per chunk), so guidance only doubles the cheap action-expert cost, not the whole model. The guidance scale $\alpha$ is tuned per garment type; in the final submission it lands in the pretty high $7$--$9$ range (\S7.7).

\subsection{Best-of-N candidate selection}

This is where the FM query's ``Q'' head (\S5.2) pays off. With $N>1$ candidates, the policy samples $N$ chunks from the \textit{same} prefix --- each with independent flow matching seed noise, so the trajectories diverge --- then scores each by the FM head's predicted $\Delta_\text{success}$ (the action-conditional gain over the value baseline, \S5.2), averaged across the two guidance passes:

\[ \text{score} = \tfrac{1}{2}\big(\Delta^\text{cond}_\text{success} + \Delta^\text{uncond}_\text{success}\big). \]

The highest-scoring chunk is executed. Candidates share the prefix, so the extra cost is only the cheap action expert.

\textbf{Retry on all-negative.} If every candidate is predicted to make the garment state worse ($\Delta_\text{success} < 0$ for all $N$), a second, larger batch is drawn and the best of the combined pool is taken.

\textbf{Which $\Delta$ to trust.} On the one hand we would want to rank by the \textit{conditional} $\Delta_\text{success}$, since it is trained on the good actions closer to the executed ones. But the conditional head is trained on positive-advantage frames only, so it is optimistically biased. The \textit{unconditional} head is unbiased over the data distribution, but it scores actions that are themselves pushed ``to the good side'' by guidance, so it is slightly mismatched. Neither score is clean, and averaging the two is the pragmatic compromise in the final solution.

It was a bit surprising this worked at all. The correlation between the FM head's prediction and the actual outcome (or its residual) was effectively zero in all my experiments --- yet 2-3-candidate rollouts consistently beat single-candidate ones. I suspect that for most chunks best-of-N does nothing: the candidates converge to the same or equally good prediction. But at the rare bottleneck states where the model is genuinely multimodal, picking the best candidate --- or avoiding the worst --- meaningfully helps.

\subsection{Initial noise: correlation and temperature}

The initial denoising noise is not i.i.d. Gaussian. It is drawn from a fitted action covariance $\mathcal{N}(0,\Sigma)$ --- with shrinkage toward the identity, $\Sigma_\text{reg} = \beta\Sigma + (1-\beta)I$ --- so it respects the joint structure of the 12-dim action chunk (also inherited from BEHAVIOR-1K~\cite{behavior1k2025}).

A noise temperature $\tau$ then scales the noise by $\sqrt{\tau}$. Values below 1 shrink the sampling variance, concentrating candidates nearer the distribution mode (\S7.7).

\subsection{Garment-type bootstrap at inference}

The policy takes the garment type as an \textit{input} (\S4.2), but at the start of an episode the runner does not yet know which garment it is looking at. Since the server keeps no state, the bootstrap lives on the client:

\begin{enumerate}
    \item A throwaway \textbf{warm-up} inference on the first observation reads the model's own predicted garment type.
    \item The first few chunks refine that estimate by \textbf{majority vote} over their predictions.
    \item The voted type is then \textbf{frozen} for the rest of the episode and fed back as the input.
\end{enumerate}

The policy also re-plans on a shorter window for those first few calls, so it reacts quickly while the garment type prediction is still settling. The late-training garment classifier is $>$99\% accurate, so the warm-up call alone is almost always right; the vote is cheap insurance. (On the real robot the client additionally holds position during the very first chunk while the type latches --- see the sim-to-real section, \S9.)

\subsection{Per-garment-type tuning via Thompson sampling}

Every knob above --- execution length, playback stretch, anchor length, inpainting onset, guidance scale, noise temperature, number of candidates --- is a hyperparameter that can also be optimized per garment type. Running a full-scale ablation and grid search over all of them would be slow, so I opted for a more efficient approach: tuning them \textbf{online, during rollout collection}, with a per-parameter Thompson-sampling bandit~\cite{thompson1933, russo2018thompson}.

\textbf{Arms and posteriors.} Each candidate value of each parameter is an arm with a $\text{Beta}(\alpha,\beta)$ posterior (e.g. the guidance scale has four arms). The parameters are optimized \textbf{independently} --- a factorized bandit, not a joint search over the full product space --- which keeps the arm count small and the posteriors well-fed.

\textbf{Sampling.} For an exploration episode, each parameter draws one sample per arm and picks the largest (standard Thompson sampling), rejecting any combination that violates $n_e + n_a \le H$. Replay and hard-mining episodes are treated as \textit{exploitation} and instead use the current best (posterior-mean) configuration.

\textbf{Reward and update.} After an episode the reward is the binary outcome with a per-type baseline subtracted, $r = \text{success} - \overline{\mathrm{SR}}(\text{type})$, added to the sampled arms as $\alpha\!\mathrel{+}=\!\max(r,0)$, $\beta\!\mathrel{+}=\!\max(-r,0)$. Only \textbf{full / random} rollouts update the bandit --- replay and curriculum episodes run on deliberately easier or harder subsets and would bias it. Each iteration the posteriors \textbf{decay} toward uniform so the bandit tracks the moving policy rather than its whole history, and the per-type posteriors are mildly \textbf{regularized} toward the pooled one.

\textbf{Freezing for submission.} Once the posteriors settle, the bandit is turned off and the best configuration per garment type is frozen for the final run. The sim-round values are in Table~\ref{tab:inference_params}:

\begin{table}[!htb]
\centering
\small
\resizebox{\columnwidth}{!}{%
\begin{tabular}{@{}lcccc@{}}
\toprule
Parameter & top\_long & top\_short & pant\_long & pant\_short \\
\midrule
Executed actions $n_e$ & 5 & 5 & 3 & 3 \\
Playback steps $\lfloor k\,n_e\rfloor$ & 5 & 5 & 3 & 3 \\
Anchor actions $n_a$ & 6 & 3 & 3 & 3 \\
Inpainting onset $t_\text{ip}$ & 0.4 & 0.4 & 0.5 & 0.5 \\
Guidance scale $\alpha$ & 7 & 7 & 9 & 7 \\
Noise temperature $\tau$ & 0.9 & 0.7 & 0.7 & 0.7 \\
Candidates $N$ & 2 & 3 & 3 & 3 \\
\bottomrule
\end{tabular}}
\caption{Frozen per-garment-type inference parameters (sim round).}
\label{tab:inference_params}
\end{table}

The values don't differ dramatically across types --- which suggests a single shared configuration would probably work nearly as well for the final model. During tuning, though, I saw individual parameters deviate significantly from the pooled values, which I attribute to the different \textit{maturity} of each garment type in training: short pants reached $>$80\% success very early and the policy shifted to optimizing for speed, while short tops were still being pushed on dexterity and accuracy and favored a different regime.

Optimizing during rollouts solves several problems at once:

\begin{itemize}
    \item a cheap way to find good hyperparameters;
    \item the hyperparameters evolve with the policy, so we stay near the best values at every stage of the RL run, not only at the end;
    \item the exploration comes for free as useful training variance --- e.g. varying the execution speed teaches the policy to complete the task faster.
\end{itemize}

Multi-armed bandits with Thompson sampling are a standard tool for optimizing the cost and convergence speed of randomized online experiments (A/B tests). Since training large policies with RL is itself expensive, I believe the same approach is a natural fit for inference-parameter optimization.

\begin{figure*}[!htb]
  \centering
  \includegraphics[width=0.85\textwidth]{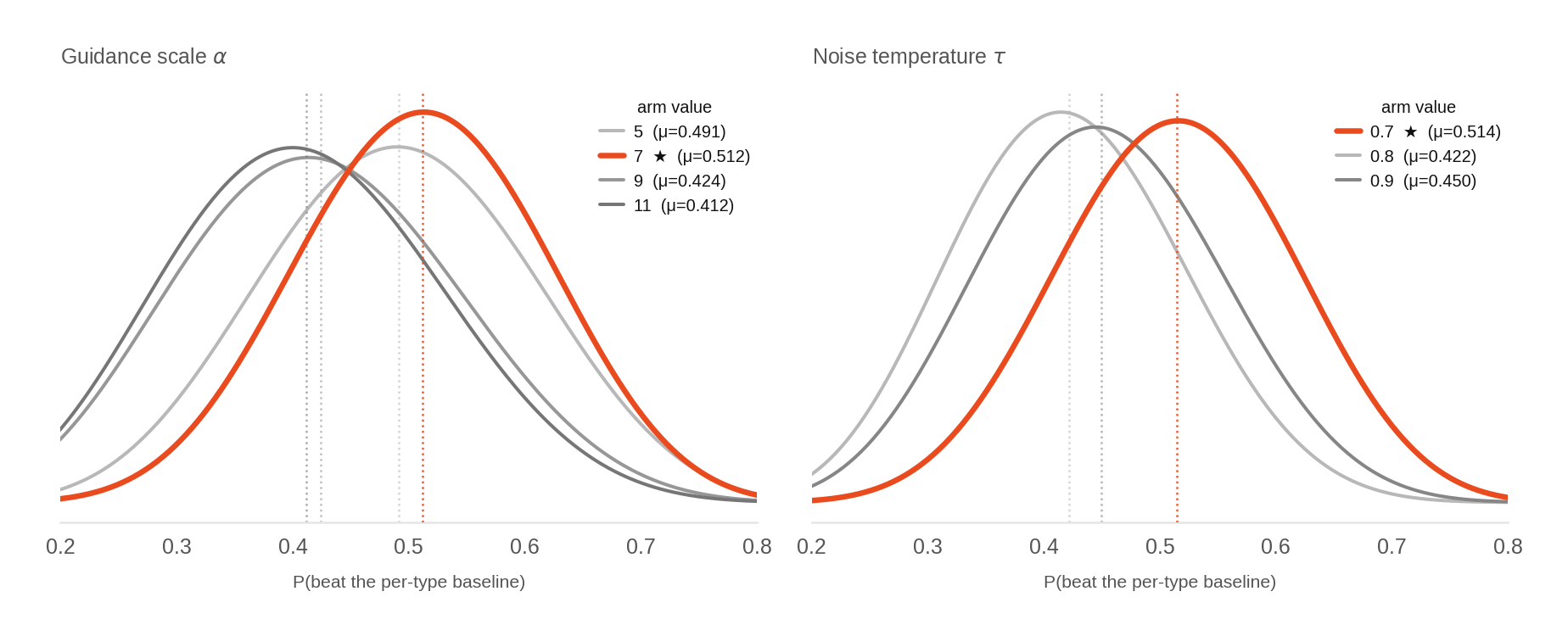}
  \caption{Thompson-sampling posteriors for short pants.}
  \label{fig:inference_bandit}
\end{figure*}

Figure~\ref{fig:inference_bandit} shows the actual converged arm posteriors for short pants late in the big run, for two of the parameters. Each curve is the $\text{Beta}(\alpha,\beta)$ posterior of one arm --- the probability that picking that value beats the per-type baseline --- so a curve shifted right is a value the bandit currently prefers. Two caveats on reading it: the reward is baseline-subtracted, so all the means sit near $0.5$ and only their \textit{relative} order matters; and the posteriors decay every iteration, so they reflect the recent on-policy window rather than the whole run. The arm sets themselves are the \textit{final} search ranges --- I adjusted them on the go as the run progressed. The guidance scale is the clearest case: I started with the conservative $0$--$2$ range, watched the posterior pin to the top arm, and repeatedly shifted the whole range upward until it settled around the $5$--$11$ window shown here. I did the same, less dramatically, for the other parameters whenever the best arm sat at an edge of the range.

The directions the bandit converged to --- even without controlled ablations --- are themselves informative about the design choices:

\begin{itemize}
    \item \textbf{Guidance scale} converged to very high values --- my initial search space was $0$--$2$, but the top of the range always dominated, so I gradually shifted it up into the $7$--$9$ region.
    \item \textbf{Number of candidates} above 1 was consistently beneficial, but going beyond 3 did not help --- consistent with best-of-N mostly \textit{avoiding very bad chunks} rather than finding an optimal one theory.
    \item \textbf{Execution length} converged to small values, suggesting the model predicts the nearest steps much more reliably than the far ones, so frequent re-planning against a fresh observation pays off.
    \item \textbf{Playback stretch} deviated from the execution length early in training but eventually matched it --- early on the model does not predict the optimal velocity, so extra exploration around the predicted speed helps; by the end it predicts the right velocity directly.
    \item \textbf{Inpainting strength} settled on light guidance: a little anchoring beats none, but strong forced inpainting limits the model's ability to correct itself and hurts performance.
\end{itemize}

\section{Online Round Results}
\label{sec:results}

The online round of the LeHome Challenge 2026 is the simulation track: the policy runs in Isaac Sim against the four garment types, each episode is scored as a binary fold success, and teams are ranked on a public leaderboard by overall success rate. The leaderboard set is 20 garments per type (80 in total): the 10 \textit{seen} garments the organizers released training data for, and 10 \textit{unseen} ones (\S1.1). Of the unseen garments, 2 per type were public --- I could use them during development, but had no organizer-provided training data for them --- and the remaining 8 per type were private, never exposed at all. My final submission took \textbf{1st place}.

I report the final leaderboard standing and a qualitative look at where the policy still fails.

\subsection{Final standing}

The online-round leaderboard ranks teams by overall success rate, averaged equally across the four garment types over all 80 garments evaluated 10 times each. 62 teams submitted. My final policy finished \textbf{1st at 79.63\%}, ahead of the second-placed team by \textbf{6.1 pp} --- a significant gap. The top of the leaderboard is shown in Table~\ref{tab:leaderboard}.

\begin{table*}[!htb]
\centering
\small
\begin{tabular}{clccccc}
\toprule
Rank & Team & Long top & Short top & Long pants & Short pants & Overall \\
\midrule
\textbf{1} & \textbf{ilya (this work)} & 74.5\% & \textbf{70.0\%} & \textbf{80.5\%} & \textbf{93.5\%} & \textbf{79.63\%} \\
2 & Shubham @ Vorwerk & 73.0\% & 62.5\% & 71.5\% & 87.0\% & 73.50\% \\
3 & Dum-E & 76.5\% & 62.0\% & 75.5\% & 79.5\% & 73.38\% \\
4 & SCUT-Unlimited & 65.5\% & 66.0\% & 70.0\% & 91.0\% & 73.13\% \\
5 & GraspYesAI & 73.5\% & 61.0\% & 69.0\% & 79.0\% & 70.63\% \\
6 & sZs & 70.5\% & 64.0\% & 68.5\% & 75.5\% & 69.63\% \\
7 & ClothFolder50k & \textbf{77.0\%} & 56.0\% & 58.5\% & 82.5\% & 68.50\% \\
8 & sisigakgak & 68.0\% & 52.5\% & 64.0\% & 77.5\% & 65.50\% \\
9 & RoboAction & 62.5\% & 61.0\% & 52.5\% & 83.0\% & 64.75\% \\
10 & MotionCrew & 64.0\% & 31.0\% & 78.5\% & 80.5\% & 63.50\% \\
\bottomrule
\end{tabular}
\caption{Top of the online-round leaderboard. (The full 62-team ranking is available on the competition website, \href{https://lehome-challenge.com/}{lehome-challenge.com}.)}
\label{tab:leaderboard}
\end{table*}

My per-type scores were 74.5\% / 70.0\% / 80.5\% / 93.5\%. The win was broad rather than carried by a single garment: I had the top score outright on short tops, long pants, and short pants, and was third on long tops (behind 77.0\% and 76.5\%). Short pants were my strongest type in absolute terms (93.5\%, against 91.0\% for the next-best on that type). Across types, short pants were the easiest for my policy and short tops the hardest --- the same ordering I saw throughout development, and one that holds across most of the leaderboard. Short tops are the most dexterity-sensitive fold (small sleeves, tight tolerances), and they were also the type the RL run was still actively pushing on at submission time. The exact checkpoint behind this result is released at \href{https://huggingface.co/IliaLarchenko/lehome_sim}{huggingface.co/IliaLarchenko/lehome\_sim}.

\subsection{Data scale and failure analysis}

The scale behind the final policy is modest by large-scale-RL standards but substantial for this competition: the final rollout dataset contained about \textbf{12,500 policy rollout episodes ($\sim$4.3M frames)} across $\sim$140 collection sessions. The pipeline continuously prunes and re-weights this pool (Sections~\ref{sec:auxheads}--\ref{sec:reward}) so that stale data decays out, and I deliberately dropped some of the oldest rollouts entirely --- the cumulative number of episodes generated over the full run was a few times larger than this retained window.

I would not call this approach sample-efficient --- I suspect the same result is reachable with a much smaller dataset. My main suspicion is the absence of any recovery logic: the initial BC dataset was scripted and had very little exploration or diversity, so it was hard for the policy to explore and find better behavior quickly.

Some qualitative observations about the remaining failures:

\begin{itemize}
    \item \textbf{Dexterity and precision.} The policy does broadly the right thing but ends up slightly off the success criteria --- the fold looks almost correct, but a keypoint distance lands just the wrong side of its threshold.
    \item \textbf{Simulation physics.} For computational stability the simulator simplifies some of the physics, which is not always perfect; this occasionally produces a failed grasp or a garment slipping out of the gripper through no real fault of the policy.
    \item \textbf{No recovery.} As noted earlier, the sim policy was not trained for recovery and saw relatively little exploration, so once it makes a mistake and lands in an out-of-distribution state it tends to fail outright rather than work its way back.
\end{itemize}

You can find some of the successful and failed rollouts in the companion blog post at \href{https://ilialarchenko.com/projects/lehome2026}{ilialarchenko.com/projects/lehome2026}.

\section{Sim-to-Real Transfer (Final Round)}
\label{sec:sim2real}

The competition had two rounds. Everything up to this point --- the policy, the RL pipeline, the inference tuning, the 1st-place online result of Section~\ref{sec:results} --- was the \textit{simulation} round. The \textit{final} round was sim-to-real: take a policy trained in Isaac Sim and make it fold real garments on a real bimanual robot.

I had roughly \textbf{three months} for the simulation round and a bit more than \textbf{one week} for sim-to-real. So the overwhelming majority of the work, experiments, and ideas in this report are about simulation; the sim-to-real part was a hackathon-style sprint where I tried to get a working transfer and fine-tune onto the real robot as quickly as possible, not a careful study. This section documents what I did and what I would do differently.

\subsection{Why zero-shot didn't work}

Simulation differs from reality in many aspects: camera sensors and placement, robot mechanics and backlash, and garment physics. The provided simulator is very capable, but the gap was still far too large for zero-shot transfer.

I am fairly confident a large part of the problem was that my policy \textbf{overfit to sim-specific details}. The clearest indirect evidence came from an image-processing test. My model resizes each 640$\times$480 camera image to 224$\times$224. If, instead of resizing directly, I first downsized to 320$\times$240 and \textit{only then} to 224$\times$224 --- a change that is nearly invisible to the human eye --- the simulation success rate dropped significantly. And when I looked at the auxiliary-head predictions, the model could \textbf{perfectly distinguish} a 640$\to$224 image from a 640$\to$320$\to$224 one. This type of overfitting doesn't transfer well to the real robot.

Some tasks were also completed differently in the real dataset than in simulation. The short-sleeve top was folded starting from the other sleeve; long pants were grabbed from the top in simulation but from the bottom (relative to the camera) in the real data; and the shorts were rotated 180 degrees in a half of the real episodes. In general you would want this kind of generalization when training a robot to fold clothes in the real world, but in the competition setting it just introduced unnecessary multimodality that hurt performance in the short run.

There was also an extra condition that made this competition harder than ordinary sim-to-real. \textbf{I never had access to the actual robot the final evaluation would run on.} So this was really \textit{sim $\to$ my robot $\to$ their robot} --- sim-to-real-to-real, with an additional generalization step baked in. Even the behavior-cloning dataset the organizers provided had been collected on a slightly different physical setup (different lighting, calibration, camera positions) than either my rig or the evaluation rig.

\begin{quote}
\textbf{A fitting illustration of how differently sim and real are perceived:} I spent two to three months certain we were folding ordinary, adult-sized clothes. It was only when I assembled the real follower rig for the final round that I realized we had been dealing with \textbf{kids' clothes} the entire time. A genuinely funny revelation --- and a good reminder that the mental model you build inside a simulator can be wrong in ways you never think to check.
\end{quote}

\subsection{Two ways to close a domain gap}

There are two fundamental levers for closing the gap between two environments: make them \textbf{more similar}, or make the training distribution \textbf{more diverse} so the gap falls inside it. I leaned on both, and the rest of this section is organized around them --- alignment work (matching cameras, calibration, units, motion intensity to the target) and diversity work (heavy augmentation, varied data sources, DAgger).

\subsection{Strategy overview}

Concretely, the sim-to-real recipe was:

\begin{enumerate}
    \item Start from a \textbf{late but not latest} checkpoint of the sim policy (\S9.4).
    \item \textbf{Strip the model} down to the heads and logic that make sense on the real robot (\S9.5).
    \item Fine-tune on a \textbf{three-bucket mix} of real organizer data, my own teleop/DAgger data, and sim replays (\S9.6).
    \item \textbf{Align} my setup and the sim as closely as possible to the target setup (\S9.7).
    \item Apply \textbf{very heavy augmentation} during training (\S9.8).
    \item \textbf{Resample each source's motion intensity} so the buckets are consistent (\S9.9).
    \item Collect \textbf{DAgger} data progressively and weight it by intervention proximity (\S9.10).
\end{enumerate}

\subsection{Starting checkpoint}

I transferred from a sim checkpoint that was strong but deliberately \textbf{not the very latest} one. The latest sim checkpoints were the most specialized to the simulator (and, per \S9.1, the most overfit to its rendering quirks); a slightly earlier checkpoint was a better-conditioned starting point for fine-tuning onto a new domain.

\subsection{Stripping the model down to the real task}

Most of the auxiliary machinery from the simulation policy depends on simulator-only privileged information (keypoint distances, success labels from the checker, advantage estimates) that does not exist on the real robot. For real training I disabled all of it and kept only what transfers:

\begin{itemize}
    \item \textbf{Still trained:} the action objective, the \textbf{garment-type} head, and the \textbf{completion} head. Garment type is needed for the inference-time bootstrap (\S7.6); completion is a progress signal that trains on any data.
    \item \textbf{Kept but frozen} (the head weights exist and load from the sim checkpoint, but get no gradient): success, mid-task checkpoint, and time-to-completion.
    \item \textbf{Removed entirely:} the keypoint-distance head and both world-model heads (the FAST-conditioned head and the action-conditional ``Q'' head) --- all of them depend on sim-only targets.
    \item \textbf{Removed from the pipeline:} advantage conditioning and the AdaRMS advantage channel (no advantage on real), and therefore classifier-free guidance and best-of-N selection at inference (both need the advantage / Q machinery). Real inference is the simpler path: a denoised chunk, soft inpainting between chunks, and the garment-type bootstrap --- nothing else.
\end{itemize}

So the real policy is essentially the initial VLM, the action expert, plus two lightweight heads, fine-tuned from a model that had learned a great deal more in sim.

\subsection{The training mix}

I fine-tuned on three buckets of data, with a fixed target share of each per training batch. Table~\ref{tab:real_data} lists the buckets, their sources, how much was collected, and the role each plays.

\begin{table*}[!htb]
\centering
\small
\begin{tabular}{@{}lrrccp{0.46\textwidth}@{}}
\toprule
Source & Episodes & Frames & Share & \shortstack{Sampling\\rate} & Role \\
\midrule
Organizer BC & 500 & 187,135 & \textbf{60\%} & $\times$1.00 & Most aligned with the eval setup (same cameras, robot colors, garments, folding pattern) --- the canonical target distribution. \\
\midrule
My home teleop & 321 & 162,363 & \multirow{2}{*}{\textbf{30\%}} & \multirow{2}{*}{$\times$0.21} & \multirow{2}{=}{Differs from the eval setup, but the only source of recovery. Kept deliberately lower because of that mismatch.} \\
My home DAgger & 471 & 288,198 & & & \\
\midrule
Sim success-replays & 1,723 & 625,111 & \textbf{10\%} & $\times$0.05 & Already-collected successful sim rollouts, replayed in a real-format pipeline with heavy augmentation --- a regularizing mix. \\
\bottomrule
\end{tabular}
\caption{The three-bucket real training mix: batch share, per-frame sampling rate, episodes, and frames per source. The two home sources (teleop + DAgger) form a single bucket and share one batch share, sampling rate, and role.}
\label{tab:real_data}
\end{table*}

The reasoning behind the weights: the organizer dataset is the closest thing I had to the actual evaluation distribution, so it dominates; my own data is valuable but collected on a mismatched rig, so it is down-weighted on purpose; sim data is there mostly to keep the model from forgetting and to add variety. Because the buckets differ so much in raw size, the fixed batch shares translate into very different per-frame sampling rates --- the large \texttt{home\_real} and \texttt{home\_sim} pools are heavily down-sampled relative to \texttt{primary\_real} (sampling-rate column of Table~\ref{tab:real_data}).

In total I collected \textbf{792 episodes ($\sim$451k frames)} of my own real-robot data on top of the organizer's 500; including the BC set, the real training pool is about \textbf{1,292 episodes / 638k frames}, with a further \textbf{1,723 sim success-replay episodes ($\sim$625k frames)} in the sim bucket.

\begin{figure}[!htb]
  \centering
  \includegraphics[width=\linewidth]{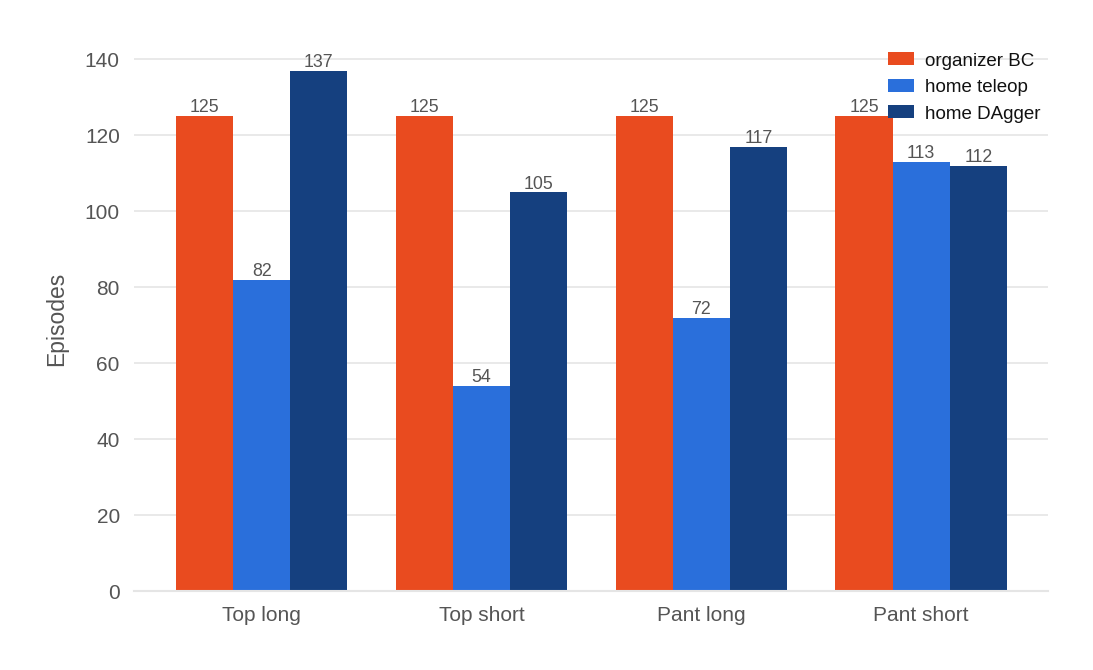}
  \caption{Real-robot episodes per garment type, split by source (organizer BC, home teleop, home DAgger).}
  \label{fig:s2r_data_overview}
\end{figure}

\begin{figure*}[!htb]
  \centering
  \includegraphics[width=0.9\textwidth]{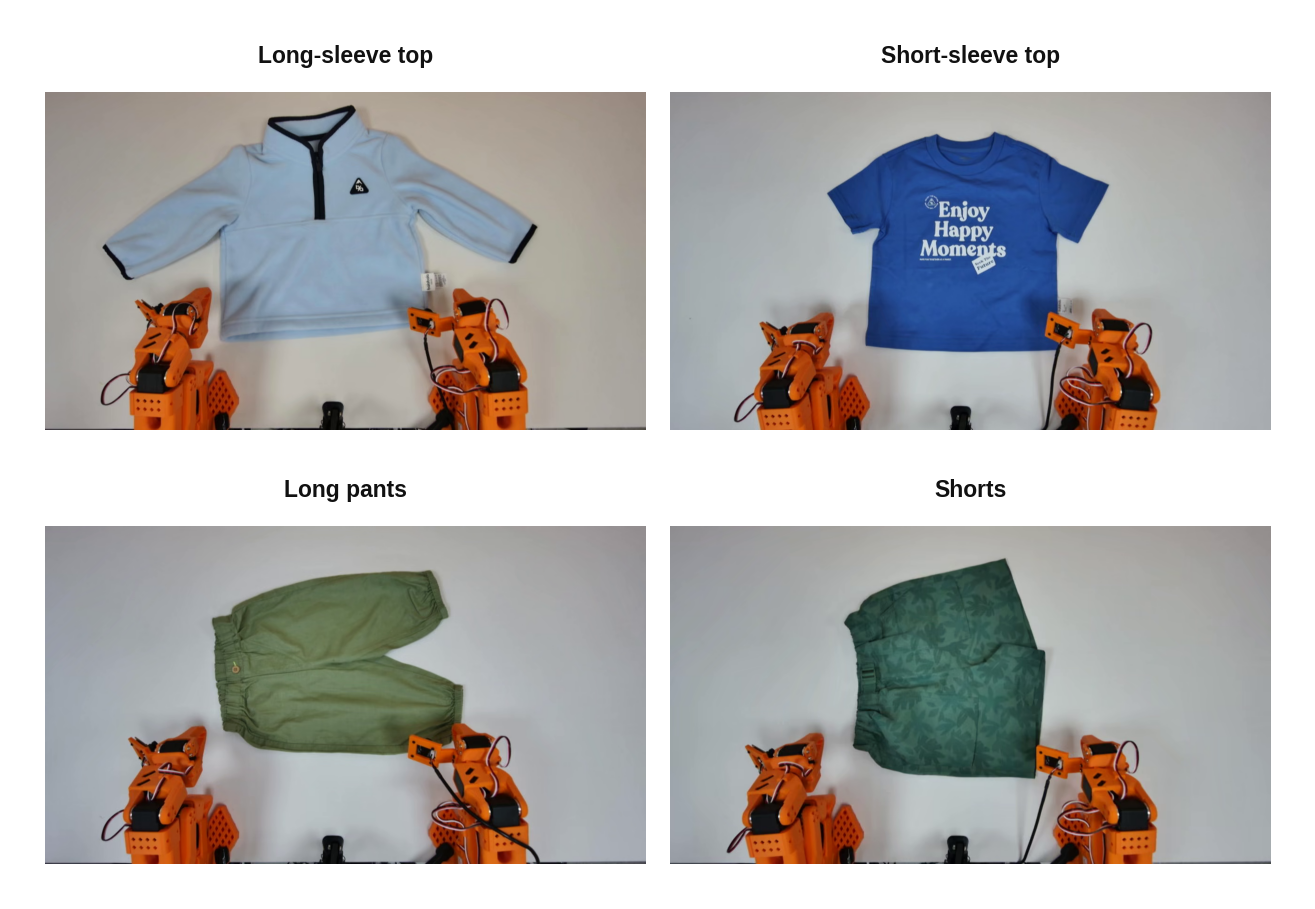}
  \caption{The four garment types in the organizer's real-robot BC dataset, seen from the overhead camera. These are the physical garments the real policy folds --- the same four types shown in their original simulation form in \S1.}
  \label{fig:s2r_garment_types}
\end{figure*}

\begin{figure*}[!htb]
  \centering
  \includegraphics[width=0.9\textwidth]{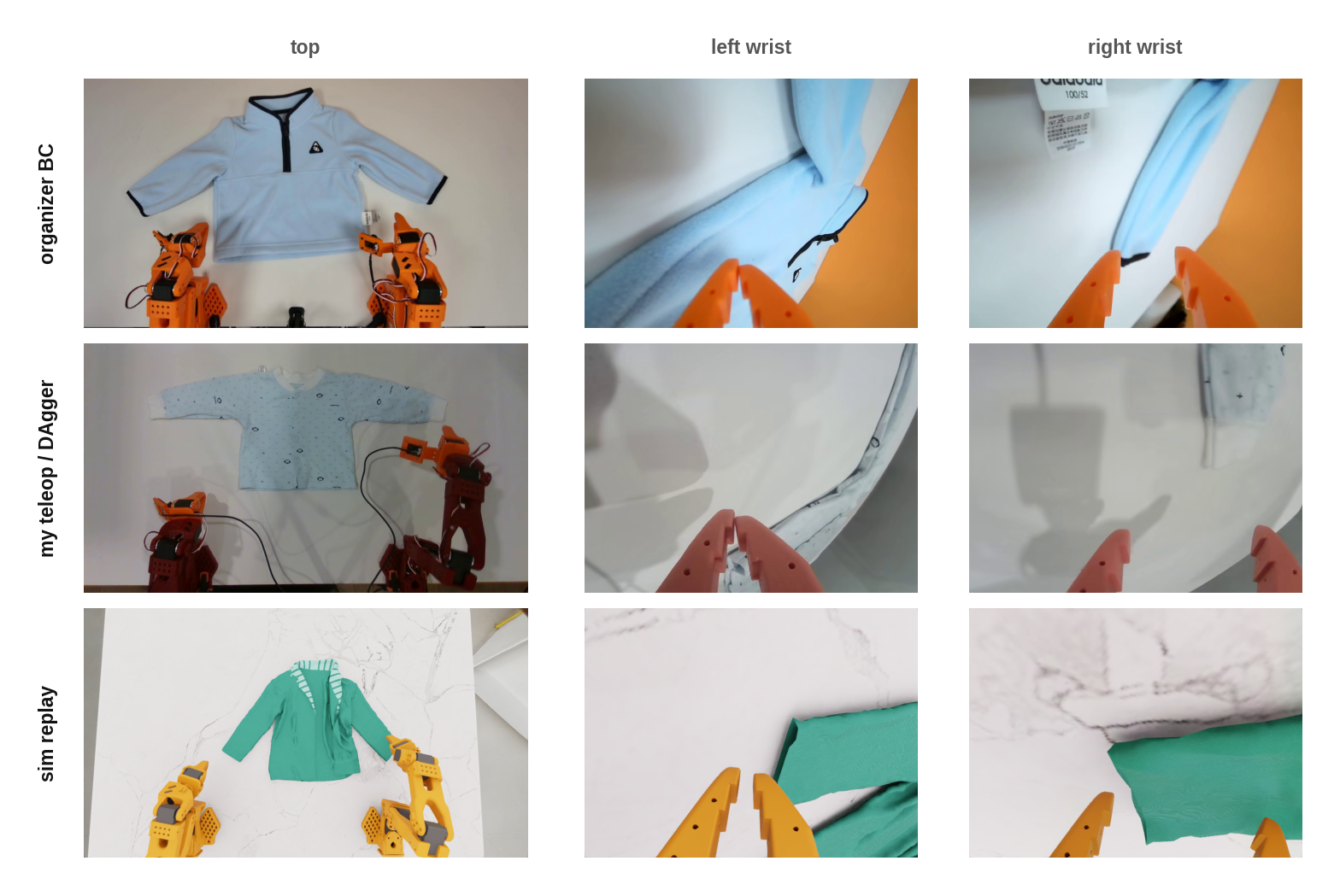}
  \caption{The three cameras (top, left wrist, right wrist) for each training bucket: the organizer BC data, my own teleop/DAgger recordings, and the sim success-replays.}
  \label{fig:s2r_dataset_montage}
\end{figure*}

Figure~\ref{fig:s2r_garment_types} shows the four physical garment types the real policy folds. The three data sources, in turn, look visibly different from each other --- different lighting, garment instances, and camera framing --- which is exactly the differences that we need to deal with (Figure~\ref{fig:s2r_dataset_montage}).

\subsection{Aligning the environments}

The alignment half of \S9.2 was mostly simple engineering: get the cameras, calibration, units, and motion to match the target as closely as I could.

\begin{figure*}[!htb]
  \centering
  \includegraphics[width=0.9\textwidth]{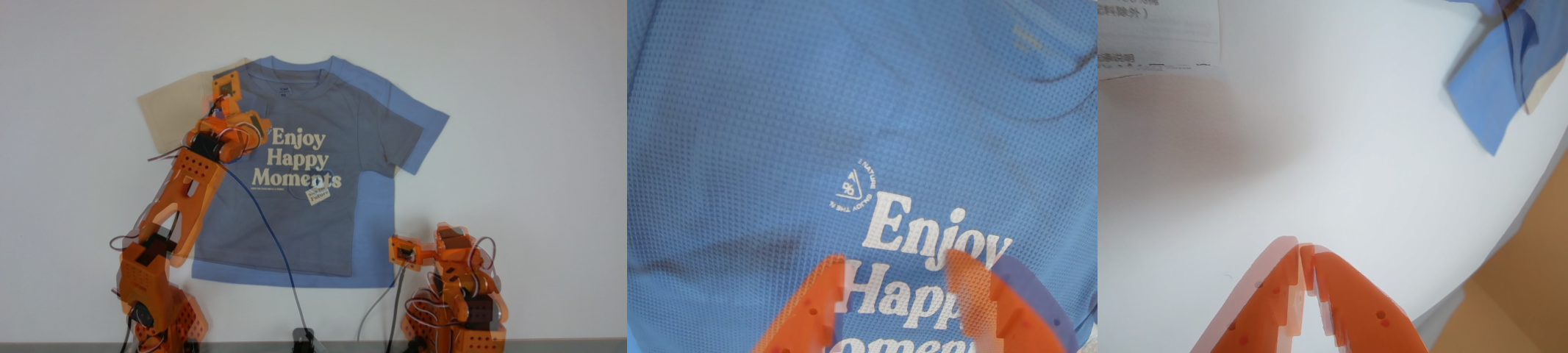}
  \caption{The three live robot cameras overlaid on the matching organizer-BC frame after driving the arms to the recorded joint state.}
  \label{fig:s2r_camera_overlay}
\end{figure*}

\begin{figure*}[!htb]
  \centering
  \includegraphics[width=0.9\textwidth]{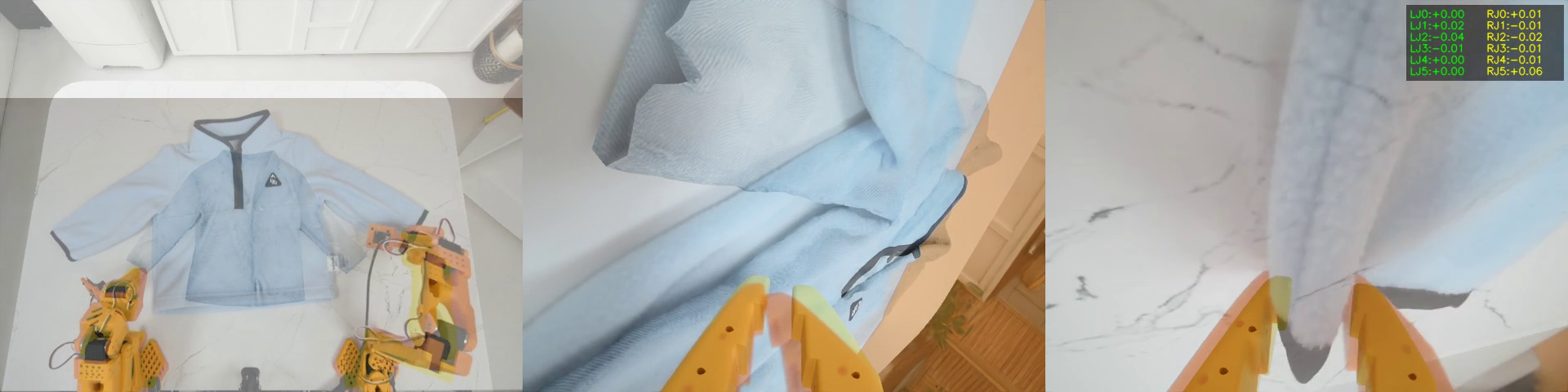}
  \caption{The same idea across the sim/real boundary: an Isaac Sim render of the same action sequence overlaid on the real cameras, used to tune the sim's camera offsets against the real rig.}
  \label{fig:s2r_sim_overlay}
\end{figure*}

The single most useful tool here was a \textbf{camera-overlay alignment} utility (Figure~\ref{fig:s2r_camera_overlay}). It picks a frame from the organizer BC dataset, drives the physical robot to that exact joint state, captures the live cameras, and overlays them on the dataset frame. When the placement is right the robot positions line up. The same idea works for overlaying a sim render against a real frame (Figure~\ref{fig:s2r_sim_overlay}). It made matching the physical rig to the source dataset a quick visual loop instead of guesswork. It turned out useful well beyond my own setup: I shared it with the organizers and other teams to help everyone align their evaluation environments at the competition.

Ideal alignment was very hard. I used different cameras, so it was never possible to overlay the images perfectly. Even during the final rollouts it seemed impossible to reproduce the evaluation setup exactly, even on the same hardware. I also deliberately did the \textit{opposite} of alignment where it helped robustness: I \textbf{randomized my own rig over time} --- moving cameras, re-calibrating the arms several times, changing the lighting --- so the model would not overfit to any one configuration. This is the diversity lever applied to the data-collection rig itself.

\subsection{Heavy augmentation}
\label{sec:heavy_aug}

\begin{figure*}[!htb]
  \centering
  \includegraphics[width=0.9\textwidth]{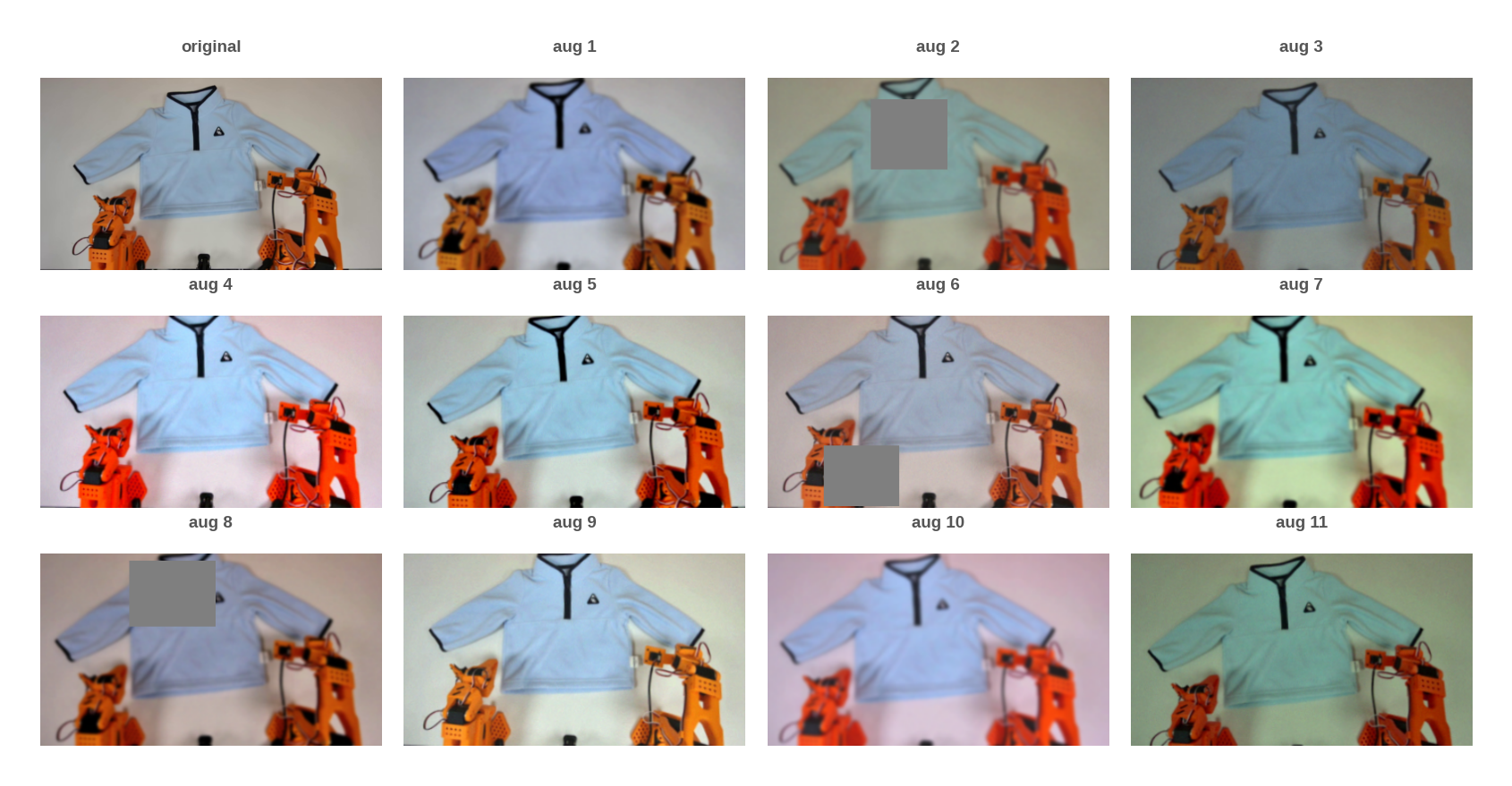}
  \caption{The same real camera frame under independent draws of the augmentation stack --- color, gain, gamma, blur, noise, crop/rotate/zoom, and cutout. Every training frame is perturbed this way so the model cannot rely on any fixed appearance.}
  \label{fig:s2r_aug_examples}
\end{figure*}

On top of environment alignment, I applied \textbf{very aggressive} training-time augmentation --- much stronger than in the sim round. The stack includes strong, per-camera-independent color jitter (the home top-camera is noticeably darker than the eval camera; sim is over-bright), per-channel gain and gamma, blur, additive sensor noise, independent crop/rotate/zoom/translate on every camera (to cover camera-mount drift), cutout, random camera dropout, and \textbf{state noise / dropout} that pushes the policy to trust the images over potentially-miscalibrated proprioception. The sim-replay bucket additionally goes through the heavy environment-augmentation engine from the data-collection round (pattern swap, color remap, camera and arm-base jitter calibrated against the real rig, table-texture and light variation). Figure~\ref{fig:s2r_aug_examples} shows the same frame under several independent draws of the stack.

\subsection{Aligning motion intensity across sources}

The three buckets were recorded at very different \textit{speeds}, and that matters because the model predicts \textbf{deltas} over a fixed horizon: a source where the arms move slowly produces small per-step deltas, a fast source produces large ones, and mixing them naively teaches the model an inconsistent notion of ``how far to move in one step.'' After RL, the sim policy moved quite fast, while my manual teleop and especially the early DAgger corrections were very slow.

So each source carries a \textbf{speed factor} that resamples its time axis before chunking, rescaling its motion intensity to match the \texttt{primary\_real} baseline (Figure~\ref{fig:s2r_speed_factors}). A factor above 1 compresses more native wall-clock time into one model step (used to ``speed up'' slow sources); a factor below 1 does the reverse (used to ``slow down'' the fast sim):

\begin{figure}[!htb]
  \centering
  \includegraphics[width=\linewidth]{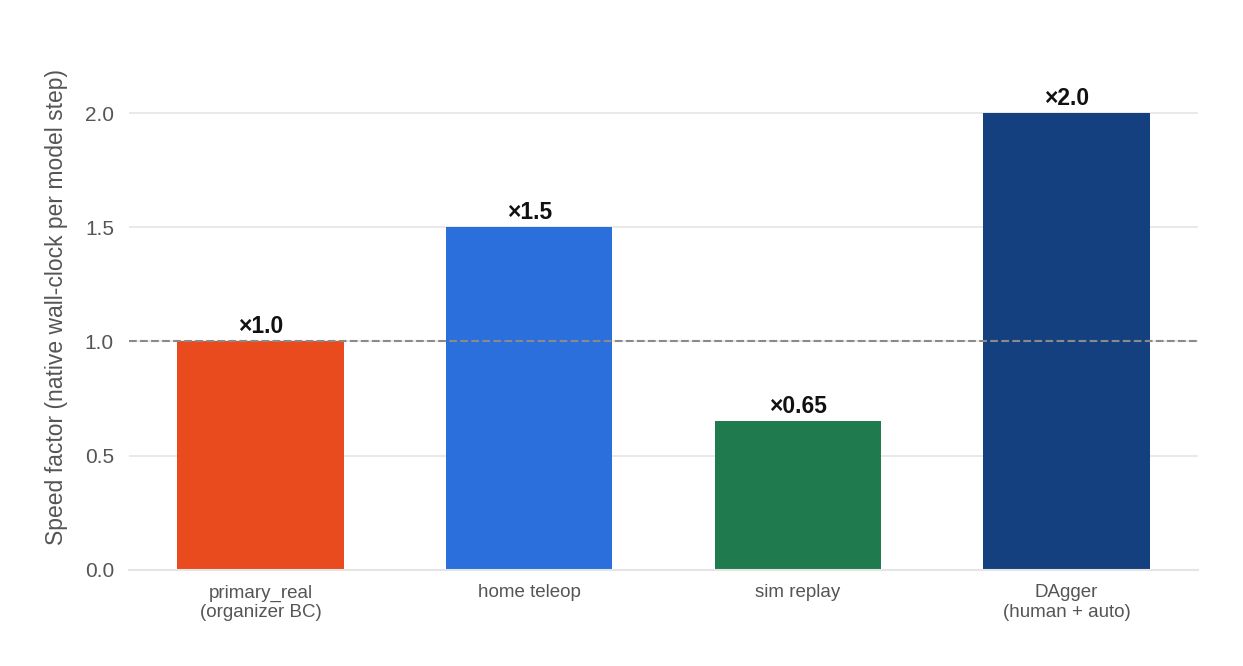}
  \caption{Per-source speed factors.}
  \label{fig:s2r_speed_factors}
\end{figure}

The organizer BC is the baseline ($\times$1.0). Home teleop is gentler and runs $\sim$1.4$\times$ longer per episode, so it is compressed ($\times$1.5). Sim rollouts are the most aggressive, so they are stretched ($\times$0.65). DAgger corrections are the slowest of all ($\sim$2$\times$ longer episodes), so they get the largest factor ($\times$2.0), and a DAgger chunk is rescaled per-chunk depending on whether its anchor frame was a human correction or an autonomous segment. As my own teleoperation got faster over the week, the right factor drifted, which is why this is set per-source rather than globally.

\subsection{DAgger collection and weighting}

Recovery data is the one thing the organizer BC set cannot provide --- it only ever shows clean, successful folds. To get off-distribution recovery behavior I ran \textbf{DAgger} on the real robot: the policy folds autonomously, and the moment it starts to fail I take over with teleop, correct it, and hand control back. Each frame is labeled by who was driving (policy vs. human).

\begin{figure}[!htb]
  \centering
  \includegraphics[width=\linewidth]{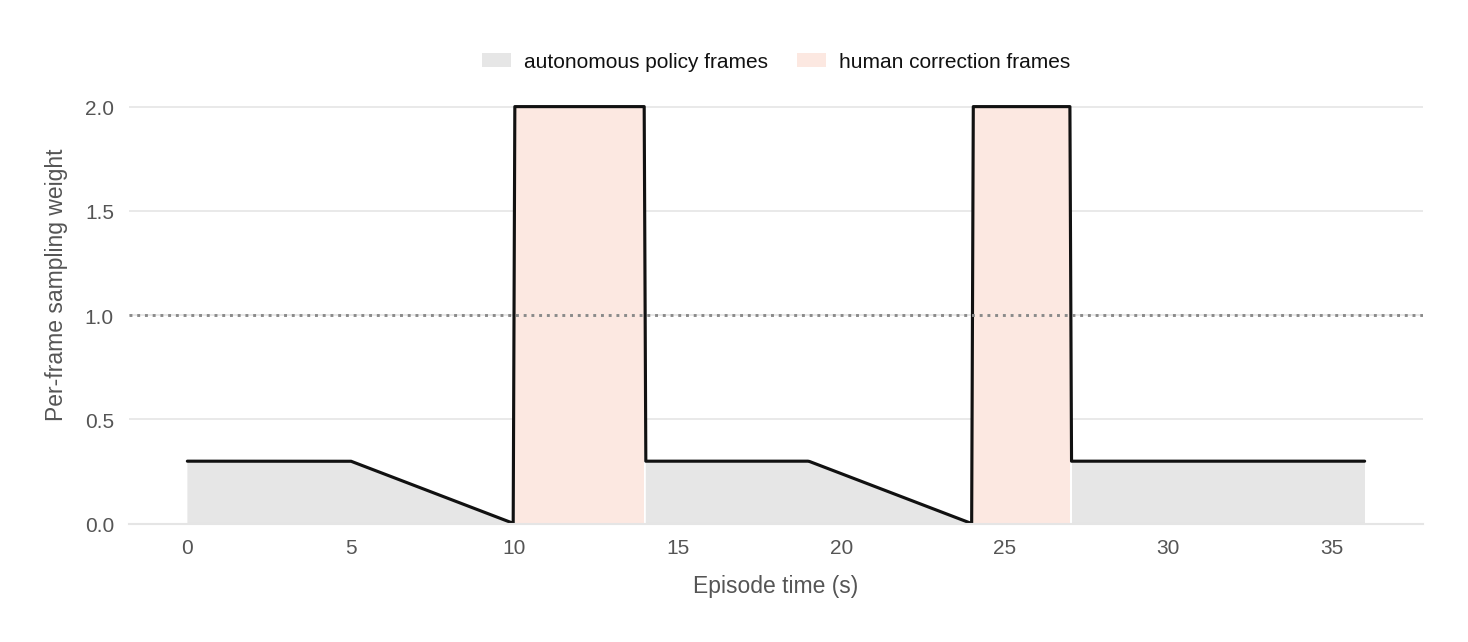}
  \caption{Per-frame sampling weight across an episode with two human interventions and a return to autonomous control. Human-correction frames get the highest weight, autonomous frames far from an intervention sit at the lower weight, and the 5-second windows just before each takeover ramp to zero.}
  \label{fig:s2r_dagger_weights}
\end{figure}

I had no reward or value function on the real robot, so I could not compute advantages the way the sim pipeline does. But the intervention signal is itself informative: a human takeover marks a \textbf{recent failure} of the policy, and the human's correction is by construction a \textbf{high-advantage} action. I turned that into per-frame sampling weights (Figure~\ref{fig:s2r_dagger_weights}):

\begin{itemize}
    \item \textbf{Human-correction frames} get the \textbf{highest} weight --- these are the demonstrations I most want the model to learn.
    \item \textbf{Autonomous frames} far from any intervention get a \textbf{low} weight (they are mostly the policy doing fine on its own --- little new signal).
    \item Autonomous frames in the \textbf{5-second window just before} a takeover \textbf{ramp down to zero}: I specifically do \textit{not} want to reinforce the exact moves that led the policy into the bad state.
\end{itemize}

This is a deliberately crude replacement for a real advantage function. A proper learned reward / value model on the real data would almost certainly help here --- both to weight frames more finely and to enable the kind of advantage conditioning and best-of-N selection that worked so well in sim --- but I simply did not have time to build and validate one in the final week.

\textbf{The collection rig.} To make DAgger ergonomic I built a setup with \textbf{two leader arms and two follower arms}, a \textbf{three-pedal foot controller} to switch quickly between control modes, everything driven from a single workstation with an RTX PRO 6000. For convenient handovers, the leader arms also track the followers while the policy is driving autonomously, so I can grab control mid-motion without a jump.

I focused DAgger collection on the \textbf{hardest cases and outliers} --- the hardest initial states where the policy tended to fail (but not too hard, the competition rules assumed flat initial garment state). In hindsight this may have been a mistake: the actual final-round initial states were mostly \textit{easier} than I expected, so effort spent on hard-state recovery was partly wasted. I suspect I would have scored higher had I instead concentrated on polishing \textbf{clean completions} of easy initial states.

\subsection{The units bug (a cautionary tale)}

I lost a lot of time to a subtle data bug. Between two LeRobot 0.4.x releases the \textbf{default state representation} for the SO-101 follower changed --- the five arm joints switched from a normalized $-100\ldots100$ range-of-motion encoding to joint angles in degrees (the gripper is a 0--100 range either way). The two conventions are related by a per-joint rescale, and for most of the joints the full range of motion is around 200 degrees, so they look almost identical on screen and the mismatch is hard to catch visually, yet it skews every joint by roughly 10\%. Most of my real data and training ran on the wrong representation before I finally tracked it down --- \textbf{only two days before the deadline}. I fixed it with a post-processing pass rather than re-collecting everything, but the disruption almost certainly cost some final performance. (The codebase has since been cleaned up so that real data is degree-mode end-to-end with no unit conversion anywhere on the training or inference path.)

\subsection{Final-round results}

I placed \textbf{2nd} in the final sim-to-real round. The top of the leaderboard, showing the teams that completed scored real-robot runs, is in Table~\ref{tab:real_results}:

\begin{table}[!htb]
\centering
\small
\begin{tabular}{@{}clc@{}}
\toprule
Rank & Team & Score \\
\midrule
1 & sZs & 895 \\
\textbf{2} & \textbf{ilya (this work)} & \textbf{865} \\
3 & Dum-E & 762.5 \\
4 & SCUT-Unlimited & 635 \\
5 & sisigakgak & 570 \\
6 & Shubham @ Vorwerk & 470 \\
\bottomrule
\end{tabular}
\caption{Top of the final sim-to-real leaderboard. Score is the organizers' composite real-robot metric, not a success-rate percentage. Both success rate and per-step quality were evaluated, unseen garments had a 50\% score bonus, and the maximum possible total was 1080 points.}
\label{tab:real_results}
\end{table}

The real-robot checkpoint behind this result is released at \href{https://huggingface.co/IliaLarchenko/lehome_real}{huggingface.co/IliaLarchenko/lehome\_real}.

\section{Discussion}
\label{sec:discussion}

The system that won the online round and came a close second in sim-to-real wasn't one clean idea --- it's a pile of practical choices made under time pressure, most of them never properly ablated. Here is what I think actually helped, what gave me trouble, and what I would do next.

\subsection{What I would keep}

\begin{itemize}
    \item \textbf{The policy as its own value function (\S5.1, \S6.2).} The success head turns the VLA into a value estimator --- no separate critic to train, sync, or serve --- and the same forward pass also gives the completion, keypoint, and $\Delta$success signals that drive advantages and best-of-N. There is still a lot of room to make these predictions more stable and better regularized, and this approach can have some downsides, so the trade offs should be assessed on a case-by-case basis.
    \item \textbf{Reward engineering (\S6).} Finding the right reward for a given task is still an art. Ground-truth checkpoints, success probability, completion, relative success --- each has trade-offs, and the trick is combining them well. My particular mix is surely far from optimal, but I think these building blocks are the right ones.
    \item \textbf{Throughput engineering (\S3.1).} Data collection, manual or automatic, is a numbers game: you have to squeeze every second. Parallelization, async failure-state generation, and anything else that saves operator time or yields more rollouts is worth it.
    \item \textbf{Inference-time tuning (\S7).} The same checkpoint can be optimized after training by tuning execution length, guidance scale, candidate count, and the rest --- found online with a cheap Thompson-sampling bandit --- was a big win that cost no extra training.
    \item \textbf{DAgger data collection.} Already standard, but worth repeating: collecting data from short human interventions is an efficient way to make a policy more robust. Whenever possible, collect in DAgger style rather than full manual teleop, and keep interventions short.
    \item \textbf{Unseen generalization was barely an issue.} I worried about the private unseen garments, but performance on them was only slightly below the trained ones, and the few misses were on garments genuinely far from anything in training. The sharpest case: a real long-pants instance that looked like shorts even to me but officially needed the long-pants fold; in one of two rollouts the policy called it short pants and stopped early --- more a labeling ambiguity than a real failure.
\end{itemize}

\subsection{What was hard or didn't work well}

\begin{itemize}
    \item \textbf{Recovery in simulation.} Teleoperating cloth through a slow sim interface is hard, and by the time the DAgger tooling worked the policy folded better than I could. Replays and augmentation made it more \textit{robust} but not better at \textit{recovering} a state already wrecked --- see \S10.4.
    \item \textbf{Overfitting to rendering artifacts.} Training only on sim, the model latched onto fine rendering and encoding details that don't transfer. It was not a problem for the sim round but required a lot of augmentation in the real one. I never found a clean fix beyond aligning the data-collection and evaluation environments, and the same overfitting also makes sim-to-real harder. 
    \item \textbf{The model is bigger than the task needs.} I used a fairly large $\pi_{0.5}$-based policy, but I think this task could be solved with a much smaller model. My main reason for entering was to experiment with full-scale VLA fine-tuning under RL, and that drove the model choice; in a more practical setting I would definitely start with something much smaller.
\end{itemize}

\subsection{Robustness to environment changes}

One thing I didn't expect was how well the policy held up to changes in the physical setup. BC policies --- especially on cheap hardware --- are usually fragile: nudge a camera a centimeter, change the focal length, shift the table, and the policy falls apart. Mine handled that kind of day-to-day drift relatively well.

I credit the deliberate rig randomization (\S9.7) and heavy augmentation (\S9.8). I varied camera pose, lighting, color, and robot-base position on purpose, so no exact geometry was ever load-bearing, while the camera-overlay alignment kept that varied distribution centered on the real rig.

More broadly, I think cheap image augmentations are underused in robotics. They are standard in computer vision and transfer almost directly to robot camera streams, yet the field spends a lot of effort on heavy neural-network image and video augmentation while the millisecond-cost classics is often unused. That has puzzled me for a while.

\subsection{Exploration and recovery: the open problem}

The hardest unsolved piece is automatic exploration. The proposed RL approach is great at reproducing and sharpening behavior it has seen, but it doesn't naturally explore its way out of states off the training distribution. Every inference-time trick I added pushes toward known-good modes, not new recovery: advantage conditioning and CFG amplify the average good behavior, best-of-N picks the least-bad of several samples from the same prefix, soft inpainting keeps the policy in the mode it already chose. None of them invent a recovery move that isn't already in the data. I experimented with a few exploration mechanisms (some traces of which are still in the code), but none of them worked well: on a flow-matching policy the extra perturbations mostly just push the chunk off the action manifold (\S2.2) and degrade performance, rather than turning up a useful new recovery move.

In practice the only reliable source of recovery was human intervention --- and it worked on the real robot but not in sim. A few seconds of real teleop correction was easy and became one of the most useful tools in the project (\S9.10). So recovery split across the two rounds: sim relied on RL and replays for robustness, real relied on DAgger for recovery. Getting a flow-matching policy to generate useful recovery attempts without a human in the loop still feels like an open problem.

\subsection{One pipeline instead of two}

My biggest regret is solving the two rounds with two mostly separate toolkits. The online round had the full RL machinery --- dense reward, value function, advantages, CFG, best-of-N. The sim-to-real round had real data and human interventions that actually produced recovery, but ran as plain BC because there was no real-side reward or value (\S9.5).

These halves are complementary. A single pipeline --- a real-side reward/value function driving advantage conditioning and best-of-N on the real robot, prioritized sampling over real rollouts, and DAgger interventions weighted by that same value signal instead of the crude hand-tuned scheme I used (\S9.10) --- would combine round 1's cleanliness with round 2's recovery. I'd also feed it more clean completions from easy starting states rather than chasing hard cases, which I over-indexed on (\S9.10). I didn't have time to build this, but I think it gets well past where either round landed alone --- my guess is 90\%+ is achievable on this task.

\subsection{Code and models}

The full training, collection, and inference stack is released at \href{https://github.com/IliaLarchenko/lehome_solution}{github.com/IliaLarchenko/lehome\_solution}. The two final checkpoints are on the Hugging Face Hub: the simulation-round policy at \href{https://huggingface.co/IliaLarchenko/lehome_sim}{huggingface.co/IliaLarchenko/lehome\_sim} and the real-world-final policy at \href{https://huggingface.co/IliaLarchenko/lehome_real}{huggingface.co/IliaLarchenko/lehome\_real}. A companion blog post with video examples is available at \href{https://ilialarchenko.com/projects/lehome2026}{ilialarchenko.com/projects/lehome2026}.


\end{document}